\documentclass[letterpaper]{article}
\usepackage{graphicx} % Required for inserting images
\usepackage{tikz-cd}
\usepackage{amssymb}
\usepackage{dsfont}
\usepackage{mleftright}
\usepackage{ntheorem}
\usepackage{mathtools}
\usepackage{algorithm, algorithmic}
\usepackage[colorlinks=true, allcolors=blue]{hyperref}
\usepackage{cleveref}
\usepackage{natbib}
\usepackage{multirow}
\usepackage{amsmath,amsfonts,graphicx}
\usepackage{xfrac}
\usepackage{tikz}
\usetikzlibrary{positioning}
\newtheorem{theorem}{Theorem}[section]
\newtheorem{lemma}[theorem]{Lemma}

\newtheorem{corollary}[theorem]{Corollary}
\newtheorem{definition}[theorem]{Definition}

\newtheorem{remark}{Remark}
\newenvironment{proof}{{\bf Proof:}}{\hfill\rule{2mm}{2mm}}

\newcommand\E{\mathbb{E}}

\renewcommand{\S}{\mathcal{S}}

\newcommand{\mail}[1]{\href{mailto:#1}{\color{blue} #1}}

\newcommand{\rr}{\mathbb{R}}

\newcommand{\calF}{\mathcal{F}}
\newcommand{\calE}{\mathcal{E}}

\newcommand{\bbE}{\mathbb{E}}

\newcommand\indicator[1]{\mathds{1}\left[ {#1} \right]}

\newcommand{\curly}[1]{ {\left\{ #1 \right\}}}
\newcommand{\roundy}[1]{ {\left( #1 \right)}}
\newcommand{\squary}[1]{ {\left[ #1 \right]}}
\newcommand{\abs}[1]{ {\left | #1 \right |}}

\newcommand{\var}{\text{Var}}

\newcommand{\norm}[1]{\left \| #1 \right \|}
\newcommand{\floor}[1]{\left \lfloor #1 \right \rfloor}

\makeatletter
\newcommand{\ForceCrefTypeInEnv}[2]{%
  \AddToHook{env/#1/begin}{%
    \let\cref@oldlabel\label
    \def\label##1{\cref@oldlabel[#2]{##1}}%
  }%
  \AddToHook{env/#1/end}{\let\label\cref@oldlabel}%
}
\makeatother

\ForceCrefTypeInEnv{lemma}{lemma}
\ForceCrefTypeInEnv{definition}{definition}
\ForceCrefTypeInEnv{proposition}{proposition}
\ForceCrefTypeInEnv{claim}{claim}
\ForceCrefTypeInEnv{corollary}{corollary}
\ForceCrefTypeInEnv{example}{example}
\crefname{lemma}{lemma}{lemmas}
\Crefname{lemma}{Lemma}{Lemmas}
\crefname{definition}{definition}{definitions}
\Crefname{definition}{Definition}{Definitions}
\crefname{proposition}{proposition}{propositions}
\Crefname{proposition}{Proposition}{Propositions}
\crefname{claim}{claim}{claims}
\Crefname{claim}{Claim}{Claims}
\crefname{corollary}{corollary}{corollaries}
\Crefname{corollary}{Corollary}{Corollaries}
\crefname{example}{example}{examples}
\Crefname{example}{Example}{Examples}
\crefname{appendix}{appendix}{appendices}
\Crefname{appendix}{Appendix}{Appendices}

\newcommand{\Qpik}{Q_k^{\pi_k}}
\newcommand{\Upik}{U_k^{\pi_k}}
\newcommand{\Upistar}{U_k^{\pi^\star}}
\newcommand{\Vpik}{V_k^{\pi_k}}
\newcommand{\Vpistar}{V_k^{\pi^\star}}
\newcommand{\hVpik}{\widehat{V}_k^{\pi_k}}
\newcommand{\hVpistar}{\widehat{V}_k^{\pi^\star}}
\newcommand{\Bpik}{B_k^{\pi_k}}

\newcommand{\Bpistar}{B_k^{\pi^\star}}
\newcommand{\qpik}{q_k^{\pi_k}}
\newcommand{\bqpik}{q_k^{\pi_k}}
\newcommand{\qpistar}{q_k^{\pi^\star}}
\newcommand{\bVpik}{\bar{V}_k^{\pi_k}}
\newcommand{\bVpistar}{\bar{V}_k^{\pi^\star}}
\newcommand{\bBpistar}{\bar{B}_k^{\pi^\star}}
\newcommand{\bBpik}{\bar{B}_k^{\pi_k}}
\newcommand{\bQpistar}{\bar{Q}_k^{\pi^\star}}
\newcommand{\Qpistar}{{Q}_k^{\pi^\star}}
\newcommand{\Vpi}{{V}^{\pi}}
\newcommand{\bVpi}{\bar{V}^{\pi}}
\newcommand{\bP}{\bar{P}}

\title{Improved Regret via Variance-Aware Policy Optimization in Adversarial MDPs}
\author{Ofir Schlisselberg \thanks{Tel Aviv University; \mail{ofirs4@mail.tau.ac.il}} \and Alon Cohen \thanks{Tel Aviv University and Google Research;
\mail{alonco@tauex.tau.ac.il}} \and Yishay Mansour \thanks{Tel Aviv University and Google Research;
\mail{mansour.yishay@gmail.com}}}

\begin{document}

\maketitle

\begin{abstract}
We study episodic adversarial MDPs with unknown dynamics and full-information rewards and achieve regret with leading term \(\widetilde O(\sqrt{H^3SAK})\). Here \(H\) is the horizon, \(S\) the total number of states across decision layers, \(A\) the number of actions, and \(K\) the number of episodes. This improves the two incomparable prior leading terms by factors \(\sqrt{S/H}\) and \(\sqrt{H^3}\), leaving a factor \(\sqrt H\) to the lower-bound scale. We give a reduction from unknown to known dynamics for \emph{downstream-dependent} algorithms, whose policy at step \(h\) is computed only from information at steps \(h+1\) and onward. This reduces the remaining leading-order gap to finding a downstream-dependent known-dynamics algorithm with \(\widetilde O(H\sqrt K)\) regret, which would yield the tight \(\widetilde O(H\sqrt{SAK})\) leading term for unknown dynamics.
We instantiate the reduction with VOPO, a policy optimization (PO) algorithm. Its improved guarantees follow from a variance-aware analysis enabled by a novel bonus, which we call the Stability-Matching Bonus (SMB). VOPO achieves \(\widetilde O(\sqrt{H^3K})\) regret under known dynamics with full information; its \(\sqrt H\) gap to the target \(H\sqrt K\) rate is inherited by our unknown-dynamics guarantee. We also show that with known dynamics and trajectory bandit feedback, VOPO achieves \(\widetilde O(\sqrt{H^2SAK})\) regret. Both known-dynamics guarantees improve the corresponding state-of-the-art PO bounds by a factor \(\sqrt H\). A matching lower bound establishes that the full-information rate is tight in one parameter regime.
\end{abstract}

\section{Introduction}

Classical reinforcement learning \citep{MannorMT-RLbook} studies MDPs with fixed transition and reward functions. Adversarial MDPs allow rewards to change adversarially across episodes. \citet{even2009online} introduced the known-dynamics setting, and \citet{neu2012adversarial} introduced the unknown-dynamics setting. \citet{zimin2013online} subsequently obtained tight regret bounds with known dynamics. In contrast, the optimal regret for unknown dynamics with adversarial full-information rewards remains unresolved, despite work under a wide range of feedback models and extensions \citep{rosenberg2019online,pmlr-v119-jin20c,lancewicki2022cooperative,jin2022delayed,lancewicki2023delay,lancewicki2025near}.

Let \(H\) be the horizon, \(S\) the total number of states across decision layers, \(A\) the number of actions, and \(K\) the number of episodes. The two best prior bounds are incomparable: \cite{rosenberg2019online} obtain \(\widetilde O(HS\sqrt{AK})\), whereas \cite{tiapkin2024narrowing} obtain \(\widetilde O(\sqrt{H^6SAK})\). Both exceed the standard \(\Omega(H\sqrt{SAK})\) lower-bound scale of \cite{domingues2021lower}.

Our algorithm achieves expected regret with leading term \(\widetilde O(\sqrt{H^3SAK})\), plus additive terms that depend on \(K\) only through logarithms. This improves the two prior leading terms by factors \(\sqrt{S/H}\) and \(\sqrt{H^3}\), respectively.\footnote{Under the loop-free convention, every decision layer is nonempty, so \(S\ge H\). The first factor is therefore at least one, with equality only when every layer has one state.} A factor \(\sqrt H\) remains to the lower-bound scale. Table~\ref{tab:unknown-dynamics-comparison} summarizes the comparison; \Cref{thm:unknown-dynamics-main} gives the full bound.

\begin{table}[t]
\centering
\small
\begin{tabular}{l c}
\hline
Result & Leading \(K\)-dependent term \\
\hline
\cite{rosenberg2019online} & \(\widetilde O(HS\sqrt{AK})\) \\
\cite{tiapkin2024narrowing} & \(\widetilde O(\sqrt{H^6SAK})\) \\
This work & \(\widetilde O(\sqrt{H^3SAK})\) \\
Standard lower-bound scale \cite{domingues2021lower} & \(H\sqrt{SAK}\) \\
\hline
\end{tabular}
\caption{Unknown dynamics with adversarial full-information rewards. Only leading terms in \(K\) are shown; logarithmic factors and additive terms are suppressed.}
\label{tab:unknown-dynamics-comparison}
\end{table}

The result follows from a reduction from unknown to known dynamics. The reduction applies to what we call \emph{downstream-dependent} known-dynamics algorithms: algorithms whose policy at step \(h\) is computed only from information at steps \(h+1\) and onward. An optimal \(\widetilde O(H\sqrt K)\) known-dynamics guarantee would yield the optimal \(\widetilde O(H\sqrt{SAK})\) leading term for unknown dynamics. We apply the reduction to an inner algorithm whose known-dynamics guarantee is a factor \(\sqrt H\) larger; this is exactly the remaining gap in our unknown-dynamics result.  Thus, the reduction both yields our improvement and identifies a concrete route to resolve the basic problem.

We construct the inner algorithm through policy optimization (PO). PO methods alternate policy evaluation with local, soft policy improvements. The standard backward computation in PO provides the downstream dependence required by the reduction. We introduce \emph{VOPO} (Variance Optimistic Policy Optimization), built around a new \emph{Stability-Matching Bonus} (SMB) and a variance-sensitive stability analysis. SMB exactly matches the stability term in these local updates, which is the key to the improved horizon dependence proved in the technical sections.

PO is also of interest independently of the reduction. It is a widely used reinforcement-learning paradigm and has demonstrated empirical success in algorithms such as TRPO and PPO \citep{schulman2015trust,schulman2017proximal}.  These successes have motivated substantial interest in its theoretical guarantees.  Accordingly, our known-dynamics PO guarantees are valuable in their own right: \(\widetilde O(\sqrt{H^3K})\) regret with full information and \(\widetilde O(\sqrt{H^2SAK})\) regret with trajectory bandit feedback. These guarantees improve the corresponding PO bounds of \cite{even2009online,luo2021policy} by a factor \(\sqrt H\), while remaining a factor \(\sqrt H\) from the best known unrestricted rates.

\paragraph{Summary of contribution.}
\begin{itemize}
    \item We improve the regret guarantee for unknown dynamics with adversarial full-information rewards, reducing the leading \(K\)-dependent term to \(\widetilde O(\sqrt{H^3SAK})\).
    \item We give a reduction from unknown dynamics to downstream-dependent known-dynamics algorithms, identifying the constrained known-dynamics problem as a route to close the remaining gap.
    \item We introduce VOPO, SMB, and a variance-sensitive analysis, obtaining \(\widetilde O(\sqrt{H^3K})\) regret with full information and \(\widetilde O(\sqrt{H^2SAK})\) regret with trajectory bandit feedback, both improving the corresponding state-of-the-art PO bounds by a factor \(\sqrt H\). In the full-information case, we give a matching lower bound, showing that our rate is tight in one regime.
\end{itemize}

\subsection{Related Work}

\paragraph{Stationary MDPs.}
For stationary MDPs with unknown dynamics, \citet{jaksch2010near} gave the classical
\(\widetilde O(DS\sqrt{AT})\) regret bound for communicating MDPs, which is
\(\widetilde O(HS\sqrt{AT})\) when the diameter is of order \(H\); the related
REGAL analysis of \citet{bartlett2009regal} replaces the diameter by a bias-span
parameter under additional information.  In finite-horizon tabular MDPs,
\citet{azar2017minimax} were the first to attain the tight first-order term
\(\widetilde O(H\sqrt{SAK})\).  Subsequent work sharpened model-free guarantees,
adaptivity, lower-order terms, or boundedness requirements
\citep{jin2018qlearning,zanette2019tighter,zhang2020reference,zhang2021mvp,
zhang2025settling,lee2025minimax}.

A common route in this literature controls transition-estimation error through
concentration for the optimal value function, after establishing a strong
connection between the policy used in an episode and the optimal policy.  This
route is unavailable for adversarial rewards, where the online-learning argument
compares the two policies only cumulatively.  \citet{zhang2025settling}
introduced \emph{Profiles} to avoid this comparison: it groups episodes by a
compact record of the data used to build their empirical models, making the
changing value functions amenable to uniform concentration.  We use Profiles to
bound the model-bias term.  \citet{lee2025minimax} introduced
\emph{quasi-optimism}, which attains the requisite control without placing a
variance estimate in the implemented bonus.  We use an adapted version of its
bonuses; \Cref{sec:unknown-dynamics} gives the relevant details.

\paragraph{Adversarial MDPs with known dynamics.}
For continuing MDPs, \citet{even2009online} gave a full-information policy-update
method with \(\widetilde O(\tau^2\sqrt K)\) regret under a uniform mixing-time
assumption; adapted to episodic MDPs, this is \(\widetilde O(H^2\sqrt K)\).
For episodic MDPs with known dynamics, \citet{zimin2013online}
introduced O-REPS, an occupancy-measure method attaining the tight \(\widetilde
O(H\sqrt K)\) with full information and \(\widetilde O(\sqrt{HSAK})\) with
trajectory bandit feedback.  Earlier, \citet{neu2010bandit} obtained
\(\widetilde O(H^2\sqrt{AK/\alpha})\) with MDP-EXP3 under the strong assumption
that every state is reached by every policy with probability at least \(\alpha\).
\citet{jinluo2020simultaneously} matched the O-REPS bandit rate with FTRL, while
\citet{wangdong2020fpl} and \citet{dai2022ftpl} gave follow-the-perturbed-leader
alternatives with \(\widetilde O(H^2\sqrt K)\) full-information and
\(\widetilde O(H\sqrt{SAK})\) bandit regret, respectively.  For aggregate
feedback with known dynamics, \citet{lancewicki2025near} gave the tight
policy-optimization bound \(\widetilde O(H^2\sqrt{SAK})\).

\paragraph{Adversarial MDPs with unknown dynamics.}
With full-information feedback, \citet{neu2012adversarial}
introduced FPOP with \(\widetilde O(HSA\sqrt K)\) regret.  UC-O-REPS of
\citet{rosenberg2019online}, and the refined FPL analysis of
\citet{wangdong2020fpl}, improved this to \(\widetilde O(HS\sqrt{AK})\).
\citet{tiapkin2024narrowing} later obtained \(\widetilde O(H^3\sqrt{SAK})\)
using policy optimization.  For unknown dynamics with trajectory bandit feedback,
the \(K^{3/4}\) OMD guarantee of \citet{rosenberg2019bandit} was improved to
\(\widetilde O(HS\sqrt{AK})\) by UOB-REPS of \citet{pmlr-v119-jin20c}.
\citet{shani2020optimistic} gave a policy-optimization alternative with a
\(K^{2/3}\) guarantee.  \citet{luo2021policy} established the first \(\widetilde O(\sqrt K)\)
policy-optimization bound, \(\widetilde O(H^2S\sqrt{AK})\), and
\citet{dai2022ftpl} matched the UOB-REPS rate with FTPL.  Finally, for aggregate
feedback, \citet{cohen2021aggregate} reduced the unknown-dynamics problem to
distorted linear bandits and obtained \(\operatorname{poly}(H,S,A)\sqrt K\)
regret; \citet{lancewicki2025near} subsequently gave the policy-optimization
bound \(\widetilde O(H^3S\sqrt{AK})\).  Here and above, \(S\) counts states
across decision layers and logarithmic factors are suppressed.

\section{Setting}

We consider an episodic, loop-free MDP with horizon \(H\), state space \(\mathcal S=\bigcup_{h=1}^H \mathcal S_h\), where \(\mathcal S_h\) is the set of states at layer \(h\), and action space \(\mathcal A\). We denote \(S=|\mathcal S|\) and \(A=|\mathcal A|\). Each episode starts at a fixed initial state \(s_1\in\mathcal S_1\). The transition function is denoted by \(P\), where \(P(s'\mid s,a)\) is the probability of transitioning from \(s\in\mathcal S_h\) to \(s'\in\mathcal S_{h+1}\) after taking action \(a\). For a function \(V:\mathcal S\to\mathbb R\), we use the shorthand

$$
PV(s,a) \coloneqq \sum_{s'\in\mathcal S}P(s'\mid s,a)V(s').
$$

Let \(\Pi\) denote the set of randomized Markov policies, where \(\pi(a\mid s)\) is the probability of choosing action \(a\) at state \(s\). An oblivious adversary assigns a reward function \(r_k:\mathcal S\times\mathcal A\to[0,1]\) to each episode \(k\in[K]\). For a policy \(\pi\), define its value and action-value functions recursively by

$$
V_k^\pi(s)
\coloneqq
\sum_{a\in\mathcal A}\pi(a\mid s)Q_k^\pi(s,a),
\qquad
Q_k^\pi(s,a)
\coloneqq
r_k(s,a)+PV_k^\pi(s,a),
$$

with the value after the horizon taken to be zero. At the initial state $s_1$, we may omit the state argument and write $V \coloneqq V(s_1)$. We denote by

$$
q^\pi(s,a)\coloneqq
\Pr_{P,\pi}(s_h=s,a_h=a),\qquad q^\pi(s)\coloneqq
\Pr_{P,\pi}(s_h=s),
\qquad s\in\mathcal S_h,  a\in \mathcal{A}
$$

the occupancy measure induced by \(P\) and \(\pi\).

At the beginning of episode \(k\), the learner selects a policy \(\pi_k\in\Pi\) and follows it in the MDP. Over \(K\) episodes, its regret against the best fixed Markov policy in hindsight is

$$
\text{Regret}_K
\coloneqq
\max_{\pi\in\Pi}\sum_{k=1}^K V_k^\pi(s_1)
-
\sum_{k=1}^K V_k^{\pi_k}(s_1).
$$

We consider three settings. In the \textbf{known-dynamics, full-information} setting, \(P\) is known and the entire reward function \(r_k\) is revealed after each episode. In the \textbf{known-dynamics, bandit-feedback} setting, \(P\) is known, but the learner observes only the rewards \(r_k(s_{k,h},a_{k,h})\) along its realized trajectory. In the \textbf{unknown-dynamics, full-information} setting, \(P\) is unknown and the entire reward function \(r_k\) is revealed after each episode.

We denote by \(\bar P_k\) the empirical transition function constructed from the observations collected through episode \(k\). A bar indicates a quantity evaluated under the empirical dynamics: \(\bar V_k^\pi\), \(\bar Q_k^\pi\), and \(\bar q_k^\pi\) denote the value function, action-value function, and occupancy measure induced by \(\bar P_k\), respectively.

\section{Stability-Matching Policy Optimization}

VOPO combines variance-sensitive Hedge updates with a bonus that exactly matches their stability term. The match shifts the stability cost from the comparator's occupancy measure to the learner's, where MDP variance bounds improve the horizon dependence.

\subsection{Variance Stability in Episodic MDPs}

We use the following refinement of Hedge at every state. Let \(x_1,\ldots,x_K\in\Delta_d\) be its uniformly initialized iterates with learning rate \(\eta\), updated using payoff vectors \(u_k\) satisfying \(0\leq \eta u_k(a)\leq 1\). For every comparator \(x^\star\in\Delta_d\),
\begin{align}\label{eq:variance_hedge}
    \sum_{k=1}^K \langle u_k,x^\star-x_k\rangle
    \leq \frac{\log d}{\eta}
    +\eta\sum_{k=1}^K\var_{a\sim x_k}\bigl(u_k(a)\bigr).
\end{align}
The stability cost is an action variance. A standard potential argument with a variance-sensitive moment-generating-function bound proves \eqref{eq:variance_hedge}; see \Cref{lem:mgf,lem:imp_hedge}.

In an MDP, occupancy-weighted action variances admit a law-of-total-variance bound. For every policy \(\pi\) and nonnegative reward function \(r\),
\begin{align}\label{eq:mdp_ltv}
    &\sum_s q^\pi(s)\var_{a\sim\pi(\cdot\mid s)}\bigl(Q^\pi(s,a)\bigr)
    +\sum_{s,a}q^\pi(s,a)\var_{s'\sim P(\cdot\mid s,a)}\bigl(V^\pi(s')\bigr) \notag\\
    &\hspace{2cm}\leq H\sum_{s,a}q^\pi(s,a)r(s,a)^2.
\end{align}
The first term has precisely the form of the stability term in \eqref{eq:variance_hedge}. For rewards in \([0,1]\), the first term sums to at most order \(H^2K\) across \(K\) episodes. Bounding a second moment using \(Q^\pi\leq H\) instead gives \(H^3K\). This difference yields the \(\sqrt{H}\) improvement after optimizing the learning rate. Equation \eqref{eq:mdp_ltv} is a direct application of the law of total variance; its proof appears in \Cref{lem:ltv}.

We will also use a second variance bound for continuation values. If \(V=\pi Q\), \(Q-PV\geq0\), and \(Q(s,a)\leq R\), then
\begin{align}\label{eq:continuation_variance}
    \sum_{s,a}q^\pi(s,a)\var_{s'\sim P(\cdot\mid s,a)}\bigl(V(s')\bigr)
    +\sum_sq^\pi(s)\var_{a\sim\pi(\cdot\mid s)}\bigl(Q(s,a)\bigr)
    \leq 2R V(s_1).
\end{align}
This continuation-value variance bound uses an upper bound on the relevant \(Q\)-functions without requiring a bound on the cumulative reward along each trajectory. It will control the bonus recursion below. Its proof appears in \Cref{lem:var_bound}.

There is, however, an obstacle to applying \eqref{eq:mdp_ltv} directly. For every comparator \(\pi^\star\), the standard value-difference identity gives
\begin{align}\label{eq:standard_vdl}
    V_k^{\pi^\star}-V_k^{\pi_k}
    =
    \sum_s q^{\pi^\star}(s)
    \sum_a\bigl(\pi^\star(a\mid s)-\pi_k(a\mid s)\bigr)Q_k^{\pi_k}(s,a).
\end{align}
Applying \eqref{eq:variance_hedge} to the local updates therefore gives the standard policy-optimization bound
\begin{align}\label{eq:standard_po_stability}
    \sum_{k=1}^K\bigl(V_k^{\pi^\star}-V_k^{\pi_k}\bigr)
    \leq \frac{H\log A}{\eta}
    +\eta\sum_{k=1}^K\sum_s q^{\pi^\star}(s)
    \var_{a\sim\pi_k(\cdot\mid s)}\bigl(Q_k^{\pi_k}(s,a)\bigr).
\end{align}
Hedge therefore produces a comparator-weighted stability term, whereas \eqref{eq:mdp_ltv} controls the learner-weighted quantity. SMB resolves this mismatch by matching the stability term to the comparator's bonus, which cancels when we return to the original rewards.

\subsection{VOPO and Stability Matching}

After each episode, VOPO receives a nonnegative payoff function: the reward function under full information or an estimator under bandit feedback. Throughout this subsection, assume
\begin{align}\label{eq:scaled_reward_assumption}
    0\leq\eta r_k(s,a)\leq \frac{1}{12H}
    \qquad\text{for all } k,s,a.
\end{align}
The condition is on the scaled payoff \(\eta r_k\), which allows the bandit estimator below as an input. VOPO evaluates \(\pi_k\) under \(r_k\) and applies a Hedge update at every state. The bonus value and action-value functions are both denoted by \(B_k\), with their arguments distinguishing them. We denote $U_k = Q_k + B_k$.

\begin{algorithm}[h!]
\caption{VOPO: Variance Optimistic Policy Optimization}
\label{alg:vopo-main}
\begin{algorithmic}[1]
\STATE \textbf{Input:} learning rate \(\eta>0\).
\STATE Set \(\pi_1(a\mid s)=1/A\) for every \(s,a\).
\FOR{\(k=1,\ldots,K\)}
    \STATE Execute \(\pi_k\) and obtain the nonnegative payoff function \(r_k\).
    \STATE Set \(V_k(s)=B_k(s)=0\) for \(s\in\mathcal S_{H+1}\).
    \FOR{\(h=H,H-1,\ldots,1\)}
        \FORALL{\(s\in\mathcal S_h\)}
            \FORALL{\(a\in\mathcal A\)}
                \STATE Set \(Q_k(s,a)=r_k(s,a)+PV_k(s,a)\).
            \ENDFOR
            \STATE Set
            \[
                b_k(s)=\eta\var_{a\sim\pi_k(\cdot\mid s)}
                \bigl(Q_k(s,a)+PB_k(s,a)\bigr).
            \]
            \FORALL{\(a\in\mathcal A\)}
                \STATE Set \(B_k(s,a)=b_k(s)+PB_k(s,a)\) and
                \(U_k(s,a)=Q_k(s,a)+B_k(s,a)\).
            \ENDFOR
            \STATE Set \(V_k(s)=\sum_a\pi_k(a\mid s)Q_k(s,a)\) and
            \(B_k(s)=\sum_a\pi_k(a\mid s)B_k(s,a)\).
        \ENDFOR
    \ENDFOR
    \FORALL{\(s\in\mathcal S,\;a\in\mathcal A\)}
        \STATE Set
        \[
        \pi_{k+1}(a\mid s)
        =
        \frac{\pi_k(a\mid s)\exp\bigl(\eta U_k(s,a)\bigr)}
        {\sum_{a'}\pi_k(a'\mid s)\exp\bigl(\eta U_k(s,a')\bigr)}.
        \]
    \ENDFOR
\ENDFOR
\end{algorithmic}
\end{algorithm}

The SMB is state-only:
\begin{align}\label{eq:smb_main}
    b_k(s)
    &=\eta\var_{a\sim\pi_k(\cdot\mid s)}\bigl(U_k(s,a)\bigr)\notag\\
    &=\eta\var_{a\sim\pi_k(\cdot\mid s)}
    \bigl(Q_k(s,a)+PB_k(s,a)\bigr).
\end{align}
The second equality holds because the current bonus \(b_k(s)\) is constant across actions and disappears from the action variance. Future bonuses enter through \(PB_k\), so the bonus is computed by backward induction without a local self-dependence.

We next show that SMB resolves the occupancy-measure mismatch. Fix a comparator \(\pi^\star\). Write \(B_k^\pi\) for the value of the bonus under \(\pi\), and \(U_k^\pi\) for its value under the augmented payoff \(r_k+b_k\), omitting the initial-state argument. The augmented counterpart of \eqref{eq:standard_vdl} gives
\begin{align}\label{eq:augmented_vdl}
    U_k^{\pi^\star}-U_k^{\pi_k}
    =
    \sum_s q^{\pi^\star}(s)
    \sum_a\bigl(\pi^\star(a\mid s)-\pi_k(a\mid s)\bigr)U_k(s,a).
\end{align}
The argument below establishes \(\eta U_k(s,a)\leq1\), as required to apply \eqref{eq:variance_hedge} at every state. By \eqref{eq:smb_main},
\begin{align}\label{eq:comparator_stability}
    \sum_{k=1}^K\bigl(U_k^{\pi^\star}-U_k^{\pi_k}\bigr)
    &\leq \frac{H\log A}{\eta}
    +\sum_{k=1}^K\sum_s q^{\pi^\star}(s)b_k(s) \notag\\
    &=\frac{H\log A}{\eta}+\sum_{k=1}^K B_k^{\pi^\star}.
\end{align}
The comparator-weighted stability cost is exactly the comparator's cumulative bonus. Converting from augmented values to the original values cancels it:
\begin{align}\label{eq:stability_cancellation}
    \sum_{k=1}^K\bigl(V_k^{\pi^\star}-V_k^{\pi_k}\bigr)
    &=\sum_{k=1}^K
    \bigl(U_k^{\pi^\star}-B_k^{\pi^\star}
    -U_k^{\pi_k}+B_k^{\pi_k}\bigr)\notag\\
    &\leq\frac{H\log A}{\eta}+\sum_{k=1}^K B_k^{\pi_k}.
\end{align}
Only the learner's bonus remains in the regret bound.

We next establish the requirement \(\eta U_k(s,a)\leq1\). The reward condition \eqref{eq:scaled_reward_assumption} directly gives
\begin{align}\label{eq:scaled_q_bound}
    \eta Q_k(s,a)\leq\frac{1}{12}.
\end{align}
A constant bound on each scaled bonus \(\eta b_k(s)\) would be insufficient: summing such bounds in the Bellman recursion would give only \(\eta B_k=O(H)\), too large to ensure \(\eta U_k(s,a)\leq1\).

The continuation-value variance bound avoids this accumulation. Given the induction hypothesis, the \(Q\)-like quantity \(Q_k+PB_k\) is bounded by \(1/(4\eta)\). Applying \eqref{eq:continuation_variance} to this quantity over the remainder of the episode yields
\begin{align*}
    B_k(s)
    \leq \frac{1}{2}\bigl(V_k(s)+B_k(s)\bigr),
    \qquad\text{and hence}\qquad
    B_k(s)\leq V_k(s).
\end{align*}
Together with \eqref{eq:scaled_q_bound}, this completes the backward induction and yields
\begin{align}\label{eq:scaled_boundedness}
    \eta Q_k(s,a)\leq\frac{1}{12},
    \qquad
    \eta B_k(s)\leq\frac{1}{12},
    \qquad
    \eta B_k(s,a)\leq\frac{1}{6}.
\end{align}
Consequently, \(\eta U_k(s,a)\leq1\), as required by Hedge. The scaled bonus value is bounded independently of the horizon. The full argument appears in \Cref{lem:Qbounded,lem:Bbounded}.

To bound the learner's bonus, combine \eqref{eq:smb_main} with \(\var(X+Y)\leq2\var(X)+2\var(Y)\):
\begin{align}\label{eq:learner_bonus_control}
    B_k^{\pi_k}
    &\leq 2\eta\sum_s q^{\pi_k}(s)
    \left[
    \var_{a\sim\pi_k(\cdot\mid s)}\bigl(Q_k(s,a)\bigr)
    +\var_{a\sim\pi_k(\cdot\mid s)}\bigl(B_k(s,a)\bigr)
    \right]\notag\\
    &\leq 2\eta H\sum_{s,a}q^{\pi_k}(s,a)r_k(s,a)^2
    +\frac{2}{3}B_k^{\pi_k}.
\end{align}
The first variance term is controlled by the standard LTV bound \eqref{eq:mdp_ltv}; the second is controlled by the continuation-value bound \eqref{eq:continuation_variance}, using \(\eta B_k(s,a)\leq1/6\). Absorbing the last term gives
\begin{align}\label{eq:learner_bonus_final}
    B_k^{\pi_k}
    \leq
    6\eta H\sum_{s,a}q^{\pi_k}(s,a)r_k(s,a)^2.
\end{align}
Thus, LTV improves the horizon dependence of the reward contribution, while the continuation-value bound controls the bonus contribution.

\begin{theorem}[Generic stability-matching guarantee]\label{thm:generic_main}
Under \eqref{eq:scaled_reward_assumption}, VOPO satisfies
\begin{align*}
    \sum_{k=1}^K\bigl(V_k^{\pi^\star}-V_k^{\pi_k}\bigr)
    \leq
    \frac{H\log A}{\eta}
    +6\eta H\sum_{k=1}^K\sum_{s,a}
    q^{\pi_k}(s,a)r_k(s,a)^2
\end{align*}
for every fixed Markov policy \(\pi^\star\).
\end{theorem}
\begin{proof}
Combine the exact cancellation \eqref{eq:stability_cancellation} with
\eqref{eq:learner_bonus_final}.
\end{proof}

Our bonuses are related to the dilated bonuses of \citet{luo2021policy}: both are state-only and control a stability term. With second-moment stability, the current bonus remains in that term, so matching it directly creates a local recursion. The dilated-bonus analysis extracts this contribution and dilates the bonus to compensate for the residual. In contrast, variance stability is invariant to the current state-only bonus, allowing SMB to match the entire stability term exactly and obtain \eqref{eq:stability_cancellation}.

\section{Known Dynamics}\label{sec:known-dynamics}

With known dynamics, VOPO can evaluate its policy and compute SMB exactly from the supplied payoff. Full information allows us to use the reward function directly; bandit feedback requires an estimator satisfying \eqref{eq:scaled_reward_assumption}.

\subsection{Full Information}

% \begin{samepage}
\begin{corollary}[Known dynamics with full information]\label{cor:full-information-known}
Suppose that \(P\) is known and that the full reward function \(r_k\in[0,1]^{\mathcal S\times\mathcal A}\) is revealed after every episode. VOPO with
$
    \eta^{-1}=\sqrt{\frac{HK}{\log A}}+12H
$
satisfies
\[
    \mathrm{Regret}_K=O\!\left(\sqrt{H^3K\log A}\right).
\]
\end{corollary}
% \end{samepage}

The learning rate ensures \(\eta r_k(s,a)\leq 1/(12H)\), so the generic guarantee gives
\[
    \mathrm{Regret}_K
    \leq \frac{H\log A}{\eta}+6\eta H^2K,
\]
using \(\sum_{s,a}q_k^{\pi_k}(s,a)=H\). Substituting the learning rate gives the claimed rate, with \(\mathrm{Regret}_K\leq HK\) covering small \(K\); see \Cref{thm:full_info_known_dynamics}.

We also prove a lower bound to complete the picture for known dynamics with full-information feedback.

\begin{theorem}[Full-information lower bound]\label{thm:full-info-known-lower-main}
For every algorithm and \(A\geq 2\), \(H\geq 1\), \(S\geq H\), and
$
    K\geq \min\{H\log A,\log(AS/H)\},
$
there is a known-dynamics MDP with full-information rewards for which
\[
    \mathrm{Regret}_K
    =
    \Omega\!\left(\sqrt{H^2K\min\{H\log A,\log(AS/H)\}}\right).
\]
\end{theorem}

At a high level, set \(n=\lfloor\min\{H/2,\log_A(S/H)\}\rfloor\). An \(n\)-step full \(A\)-ary tree maps each of its \(A^n\) action sequences to a distinct state, and distributing each leaf's expert payoff over the remaining \(H-n\) layers embeds an \(A^n\)-expert problem with payoff scale \(H-n=\Omega(H)\); the standard experts lower bound then yields the result.

\citet{zimin2013online} obtain \(O(\sqrt{H^2K\log(SA/H)})\) with O-REPS, which matches \Cref{thm:full-info-known-lower-main} when \(\log(SA/H)\leq H\log A\). VOPO matches the lower bound in the complementary regime. Thus, O-REPS and VOPO together complete the picture for known dynamics with full-information feedback. The proof of the lower bound appears in \Cref{thm:full-info-known-lower}.

The lower bound raises a structural question for PO: how can per-state policy updates attain the logarithmic state dependence suggested by \Cref{thm:full-info-known-lower-main} while removing the remaining \(\sqrt H\) factor? We return to this question in the discussion.

\subsection{Bandit Feedback}

Under bandit feedback, let \(\mathds{1}_k[s,a]\) indicate that the trajectory in episode \(k\) visits \((s,a)\). We run VOPO on the payoff estimator
\begin{align}\label{eq:known-bandit-estimator}
    \widehat r_k(s,a)
    =
    \frac{\mathds{1}_k[s,a]r_k(s,a)+2\gamma}
    {q_k^{\pi_k}(s,a)+\gamma},
    \qquad
    \gamma=36H\eta.
\end{align}
Known dynamics make \(q_k^{\pi_k}(s,a)\) available before episode \(k\). For \(\delta\in(0,1/3)\), set
\[
    \eta^{-1}=\sqrt{\frac{SAK}{\log(SA/\delta)}}+36H.
\]
The choice ensures \(\gamma<1\) and
\begin{align}\label{eq:known-bandit-properties}
    0\leq\eta\widehat r_k(s,a)\leq\frac{1}{12H}.
\end{align}
Thus \(\widehat r_k\) is a valid input to the generic guarantee.

Implicit exploration (IX), introduced for loss estimators by \citet{neu2015explore}, provides a one-sided high-probability comparison for losses. As a technical contribution of our bandit analysis, we prove a reward-side IX lemma tailored to \eqref{eq:known-bandit-estimator}. Specializing \Cref{lem:ix} and taking a union bound gives, simultaneously for every \((s,a)\),
\begin{align}\label{eq:known-bandit-ix}
    \sum_{k=1}^K r_k(s,a)
    \leq
    \sum_{k=1}^K\widehat r_k(s,a)
    +\frac{\log(SA/\delta)}{\gamma}
\end{align}
with probability at least \(1-\delta\). Thus:
\begin{align}\label{eq:comparator_ix}
    \sum_k \Vpistar &= \sum_{k,s,a}\qpistar(s,a)r_k(s,a) \le \sum_{k,s,a}\qpistar(s,a)\widehat{r}_k(s,a) + \tilde{O}\roundy{\frac{H}{\gamma}} = \sum_k \hVpistar + \tilde{O}\roundy{\frac{H}{\gamma}}
\end{align}
This is the comparator transfer needed for a high-probability regret guarantee.

\begin{theorem}[Known dynamics with bandit feedback]\label{thm:known-dynamics-bandit}
Suppose that \(P\) is known and that the learner observes only the rewards along its trajectory. With the estimator \eqref{eq:known-bandit-estimator} and the preceding choice of \(\eta\), with probability at least \(1-3\delta\), VOPO satisfies
\[
    \mathrm{Regret}_K
    =O\!\left(\sqrt{H^2SAK\log(SA/\delta)}\right).
\]
\end{theorem}

Standard concentration bounds yield the learner-side transfer
\begin{align*}
    \sum_k \hVpik
    &\leq
    \sum_k \Vpik + \widetilde O\roundy{SAK\gamma}.
\end{align*}
Together with \eqref{eq:comparator_ix}, this gives
\begin{align*}
    \mathrm{Regret}_K
    &=
    \sum_k \roundy{\Vpistar-\Vpik}
    \leq
    \sum_k \roundy{\hVpistar-\hVpik}
    +\widetilde O\roundy{SAK\gamma+\frac{H}{\gamma}}.
\end{align*}

We use \Cref{thm:generic_main} to bound the first term. It remains to control its second-moment contribution:
\begin{align*}
    \eta H\sum_{k,s,a}\qpik(s,a)\widehat{r}_k^2(s,a)
    &=
    O\roundy{\eta H\sum_{k,s,a}\widehat{r}_k(s,a)}
    =
    \widetilde O\roundy{\eta HSAK}.
\end{align*}
The first equality uses \(\qpik(s,a)\widehat{r}_k(s,a)=O(1)\); standard concentration bounds give the final estimate, up to lower-order terms. Thus, \Cref{thm:generic_main} gives
\begin{align*}
    \sum_k \roundy{\hVpistar-\hVpik}
    &=
    \widetilde O\roundy{\frac{H}{\eta}+SAKH\eta}.
\end{align*}
Using \(\gamma=O(H\eta)\) and substituting the choice of \(\eta\) proves the theorem.
The complete proof appears in \Cref{apx:bandit_known}.

\section{Unknown Dynamics}\label{sec:unknown-dynamics}

For unknown dynamics with full-information rewards, we run VOPO on an empirical model and restart it over count-based epochs. Let \(i_k\) be the epoch containing episode \(k\), and let \(N_i(s,a)\) be the number of samples of \((s,a)\) used in epoch \(i\). Here we index the empirical model by epoch, writing \(\bar P_i\) for the model constructed from these samples. Each \(N_i(s,a)\) is a dyadic prefix count. When a visit count doubles, the reduction updates the empirical model and restarts VOPO. There are at most \(SA\log K\) epochs.

\begin{algorithm}[H]
\caption{Informal empirical-model reduction; the formal version is \Cref{alg:known-to-unknown}.}
\label{alg:unknown-main}
\begin{algorithmic}[1]
\STATE Initialize an empirical model and a downstream-dependent known-dynamics inner algorithm.
\FOR{\(k=1,\ldots,K\)}
    \STATE Run the inner algorithm on \(\bar P_{i_k}\) with payoff \(r_k+b_{i_k}\), and execute its policy in the true MDP.
    \STATE Observe the trajectory and update the visit counts.
    \IF{a visit count reaches its next dyadic threshold}
        \STATE Construct the next empirical model from the corresponding dyadic sample prefixes and restart the inner algorithm.
    \ENDIF
\ENDFOR
\end{algorithmic}
\end{algorithm}

The augmented payoff uses the state-action bonus:
\begin{align}\label{eq:unknown-exploration-bonus}
    b_i(s,a)
    =
    \min\!\left\{
        \frac{\beta}{\max\{N_i(s,a),1\}},\,H
    \right\},
\end{align}
where \(\beta>0\) is specified below. This bonus compensates for uncertainty in the empirical dynamics; the inner algorithm still uses SMB to match Hedge's stability term. Its cap is essential for the reduction - since \(r_k+b_i\leq H+1\), choosing
\[
    \eta^{-1}\geq 12H(H+1)
\]
ensures \eqref{eq:scaled_reward_assumption}, allowing VOPO to use the augmented payoff without normalization.

For a policy \(\pi\), write \(\bar B_k^\pi\) for its value under the exploration bonus \(b_{i_k}\) and model \(\bar P_{i_k}\), and \(\bar V_k^\pi\) for its value under reward \(r_k\) and the same model. The regret decomposes into four terms:
\begin{align}\label{eq:unknown-regret-decomposition}
    \sum_{k=1}^K
    \bigl(V_k^{\pi^\star}-V_k^{\pi_k}\bigr)
    &=
    \underbrace{\sum_{k=1}^K
    \bigl(V_k^{\pi^\star}-\bar V_k^{\pi^\star}-\bar B_k^{\pi^\star}\bigr)}
    _{\text{quasi-optimism}}
    +
    \underbrace{\sum_{k=1}^K
    \bigl(\bar V_k^{\pi^\star}+\bar B_k^{\pi^\star}
    -\bar V_k^{\pi_k}-\bar B_k^{\pi_k}\bigr)}
    _{\text{restarted inner regret}}
    \notag\\
    &\quad+
    \underbrace{\sum_{k=1}^K\bar B_k^{\pi_k}}
    _{\text{exploration bonus}}
    +
    \underbrace{\sum_{k=1}^K
    \bigl(\bar V_k^{\pi_k}-V_k^{\pi_k}\bigr)}
    _{\text{model bias}}.
\end{align}

The first term is controlled by quasi-optimism. On the good event in the appendix, for every episode \(k\),
\begin{align}\label{eq:quasi-optimism-main}
    V_k^{\pi^\star}
    -\bar V_k^{\pi^\star}
    -\bar B_k^{\pi^\star}
    \leq
    \frac{6H^2\log(SAK)}{\beta}.
\end{align}
The important feature of \eqref{eq:quasi-optimism-main} is that the implemented bonus \eqref{eq:unknown-exploration-bonus} contains no variance estimate, yet its horizon dependence is controlled by comparator-value variance. A direct variance-aware bonus would need to estimate \(\var_{s'\sim P(\cdot\mid s,a)}(V_k^{\pi^\star}(s'))\). The usual replacement by a variance based on \(\bar V_k^{\pi_k}\) would in turn require an episodewise, statewise bound on \(V_k^{\pi^\star}(s)-\bar V_k^{\pi_k}(s)\). Such a bound is unavailable in the adversarial setting: the online optimization argument relates the learner to the comparator only after summing over episodes. The proof instead uses the auxiliary reward
\[
    c_k(s,a)
    =
    \frac{3\log(SAK)}{\beta}
    \var_{s'\sim P(\cdot\mid s,a)}
    \bigl(V_k^{\pi^\star}(s')\bigr)
\]
only in the analysis. Backward induction controls the quasi-optimism error in terms of the value of \(c_k\), and LTV gives the \(H^2/\beta\) scale. The induction and concentration calculations appear in \Cref{lem:quasi_opt}.

Restarting VOPO bounds the second term by
\[
    O\!\left(
        \frac{HSA\log K\log A}{\eta}
        +\eta H^2K
    \right),
\]
plus a contribution bounded by the cumulative exploration bonus. The latter is \(O(SA\beta\log K)\), controlling both this contribution and the third term; see \Cref{lem:reg,lem:bonus}.

The fourth term is controlled by the Profiles trick \citep{zhang2025settling}. This argument requires downstream dependence: for every episode \(k\), layer \(h\), and \(s\in\mathcal S_h\), \(\bar V_k^{\pi_k}(s)\) must be measurable with respect to the epoch sample counts through episode \(k\) and the empirical transition models at layers \(h'\geq h\). It also requires the dyadic sample prefixes described above. VOPO's backward evaluation under \(\bar P_{i_k}\) satisfies this interface. The Profiles argument then yields
\begin{align}\label{eq:profiles-bias-main}
    \sum_{k=1}^K
    \bigl(\bar V_k^{\pi_k}-V_k^{\pi_k}\bigr)
    =
    O\!\left(
        \sqrt{H^2SAK\log^3(SAK)}
        +H^2S^2A\log^2(SAK)
    \right)
\end{align}
on its good event. The Profiles proof appears in \Cref{apx:profiles}.

% \begin{samepage}
\begin{theorem}[Unknown dynamics with full information]\label{thm:unknown-dynamics-main}
Suppose that \(P\) is unknown and that the full reward function is revealed after every episode. Run the empirical-model reduction with VOPO as its inner algorithm, and set
\[
    \eta^{-1}
    =
    \sqrt{\frac{HK}{SA\log A\log K}}
    +12H(H+1),
    \qquad
    \beta
    =
    \sqrt{\frac{H^2K}{SA}}
    +48H^3S\log(SAK).
\]
Then
\[
    \mathbb E[\mathrm{Regret}_K]
    =
    \widetilde O\!\left(
        \sqrt{H^3SAK}
        +H^3S^2A
    \right).
\]
\end{theorem}

% \end{samepage}

The proof appears in \Cref{thm:full_info_unknown_dynamics}.

\citet{tiapkin2024narrowing} also use restarts and Profiles, obtaining the leading rate \(\widetilde O(\sqrt{H^6SAK})\). Our improvement by a factor \(\sqrt{H^3}\) has two sources. A factor \(\sqrt H\) comes from VOPO's variance-sensitive inner analysis. The remaining factor \(H\) comes from handling augmented payoffs as large as \(H+1\). Normalizing these payoffs to \([0,1]\) multiplies the regret by \(H\). VOPO instead uses \(\eta(r_k+b_{i_k})\leq 1/(12H)\): the cap imposes \(\eta^{-1}\geq 12H(H+1)\) without this leading-order cost.

The exploration-bonus contribution also matters for closing the remaining gap. In \citet{tiapkin2024narrowing}, it is of order \(\widetilde O(\sqrt{H^3SAK})\), which prevents an \(\widetilde O(\sqrt{H^2SAK})\) leading rate even with an optimal inner algorithm. Our quasi-optimistic bonus avoids this obstruction.

\section{Discussion}\label{sec:discussion}

The natural direction for future work is to design a downstream-dependent policy-optimization algorithm with the tight \(\widetilde O(H\sqrt K)\) regret guarantee in the known-dynamics full-information setting. Combined with our reduction, such an algorithm would yield the tight \(\widetilde O(H\sqrt{SAK})\) leading term for unknown dynamics with full-information rewards. It would therefore resolve both the tight horizon dependence for policy optimization and the unknown-dynamics full-information problem. FTPL is another promising direction, since it is downstream-dependent; obtaining an appropriate regret guarantee for FTPL could make it suitable as the inner algorithm in the reduction.

The full-information lower bound suggests that removing the \(\sqrt H\) factor requires a logarithmic dependence on the number of states. This is not immediate from the PO construction, whose local updates are defined separately at each state. A second obstacle arises for shared-rate OMD-based PO. Consider an MDP whose first \(H/2\) steps form \(H/2\) independent expert problems and whose remaining \(H/2\) steps form a single expert problem with payoff scale \(H\). Up to logarithmic factors, standard OMD lower-bound instances yield a contribution of order \(H/\eta\) from the first part and \(\eta H^2K\) from the second part. A common learning rate therefore gives
\[
    \frac{H}{\eta}+\eta H^2K
    = \Omega\!\left(\sqrt{H^3K}\right).
\]
Thus, attaining an \(O(\sqrt{H^2K})\) PO guarantee appears to require data adaptivity, such as state-dependent learning rates, rather than a shared-rate local update.

\section*{Acknowledgements}
OS and YM are supported by the European Research Council (ERC) under the European Union’s Horizon 2020 research and innovation program (grant agreement No. 882396), by the Israel Science Foundation (1357/24), the Blavatnik foundation at Tel Aviv University and by a grant from the Tel Aviv University Center for AI and Data Science (TAD). OS is also supported by the TAD Excellence Program for Doctoral Students in Artificial Intelligence and Data Science from the Tel Aviv University Center for AI and Data Science (TAD) and from the Israeli Council for Higher Education (CHE) Fellowship for Outstanding PhD Students in Data Science. AC  is supported by the Israel Science Foundation
(ISF, grant number 2250/22).

\newpage
\bibliography{refs}
\bibliographystyle{plainnat}
\newpage

\appendix
\crefalias{section}{appendix}

\section{General Lemmas}
\subsection{Value Difference Lemma}
\begin{lemma}[Value Difference Lemma]\label{lem:vdl}
For every two policies $\pi,\pi'$,
\begin{align*}
    V^\pi - V^{\pi'} &= \sum_{s}q^\pi(s)\sum_a \roundy{\pi(a \mid s) - \pi'(a\mid s)}Q^{\pi'}(s,a)
\end{align*}
\end{lemma}
\begin{proof}
For every $s,a$:
\begin{align*}
    Q^\pi(s,a) - Q^{\pi'}(s,a) &= PV^\pi(s,a) - PV^{\pi'}(s,a)
\end{align*}

Fix some $h$. For every $s\in \S_h$:
\begin{align*}
    V^{\pi}(s) - V^{\pi'}(s) &= \sum_{a}\pi(a\mid s)Q^\pi(s,a) - \pi'(a\mid s)Q^{\pi'}(s,a)\\
    &= \sum_a \pi(a\mid s)\roundy{Q^\pi(s,a) - Q^{\pi'}(s,a)} +  \roundy{\pi(a\mid s) - \pi'(a\mid s)}Q^{\pi'}(s,a)\\
    &= \sum_a \pi(a\mid s)\roundy{\sum_{s'\in \S_{h+1}}{P(s'\mid s,a)\roundy{V^\pi(s') - V^{\pi'}(s')}}} +  \roundy{\pi(a\mid s) - \pi'(a\mid s)}Q^{\pi'}(s,a)
\end{align*}

In the first term we sum over all options for $a$ and $s'\in \S_{h+1}$, thus:
\begin{align*}
    \sum_{s\in \S_h}q^\pi(s)\roundy{V^{\pi}(s) - V^{\pi'}(s)} &= \sum_{s'\in S_{h+1}}q^\pi(s')\roundy{V^{\pi}(s) - V^{\pi'}(s)} + \sum_{s\in \S_h}q^\pi(s)\sum_a \roundy{\pi(a\mid s) - \pi'(a\mid s)}Q^{\pi'}(s,a)
\end{align*}

We have a sequence of the structure $a_h = a_{h+1} + b_h$, which means $a_1 = \sum_h b_h$, thus:
\begin{align*}
    V^\pi - V^{\pi'} &= \sum_{s} q^\pi(s)\sum_a \roundy{\pi(a\mid s) - \pi'(a\mid s)}Q^{\pi'}(s,a)
\end{align*}
\end{proof}

% \begin{lemma}\label{lem:evdl}
% For every two dynamics $P,P'$ let $V,V'$ be the corresponding value function w.r.t to some policy $\pi$. We have:
% \begin{align*}
%     V - V' &= \sum_{s,a}q(s,a)\roundy{P-P'}V'(s,a)
% \end{align*}
% where $q$ is the occupancy measure w.r.t $P$.
% \end{lemma}
% \begin{proof}
% We will prove in backwards induction that:
% \begin{align*}
%     V(s) - V'(s) &= \sum_{s',a}q(s',a \mid s)\roundy{P-P'}V'(s',a)
% \end{align*}
% It is true  for every $s\in \S_{H}$ because for every $s\in \S_{H+1}$ we have $V'(s) = 0$ and
% \begin{align*}
%     V(s) = V'(s) = \sum_a \pi(a\mid s)r(s,a).
% \end{align*}

% For every $s,a$
% \begin{align*}
%     Q(s,a) - Q'(s,a) &= PV(s,a) - P'V'(s,a)\\
%     &= P(V - V')(s,a) + (P-P')V'(s,a)\\
%     &= \sum_{s'}P(s'\mid s,a)(V(s') - V'(s')) + (P-P')V'(s,a)\\
%     &= \sum_{s'} P(s'\mid s,a)\roundy{\sum_{s'', a''}q(s'', a'' \mid s')\roundy{P-P'}V'(s'',a'')} + (P-P')V'(s,a)\\
%     &= \sum_{s'\ne s,a'}q(s',a'\mid s,a)\roundy{P-P'}V'(s',a') + (P-P')V'(s,a)\\
%     &= \sum_{s',a'}q(s',a'\mid s,a)\roundy{P-P'}V'(s',a')\\
%     \implies V(s) - V'(s) &= \sum_a \pi(a\mid s)\roundy{\sum_{s',a'}q(s',a'\mid s,a)\roundy{P-P'}V'(s',a')}\\
%     &= \sum_{s',a'}q(s',a'\mid s)\roundy{P-P'}V'(s',a')
% \end{align*}
% \end{proof}

\subsection{Variance Handling}
\begin{lemma}\label{lem:var_bound}
Fix policy $\pi$ and some $R > 0$. Assume $Q(s,a) \le R$ for all $s,a$. Then:
\begin{align*}
    \sum_{s,a}q^\pi(s,a)\var_{s'\sim P(s,a)}\roundy{V^\pi(s')} + \sum_{s}q^\pi(s)\var_{a\sim\pi}\roundy{Q^\pi(s,a)} \le 2RV(s_1)
\end{align*}
\end{lemma}
\begin{proof}
We short $V\coloneqq V^\pi$, $Q\coloneqq Q^\pi$, $q \coloneqq q^\pi$. 

For every $s$:
\begin{align*}
    \var_{a\sim\pi}\roundy{Q(s,a)} &= \pi Q^2(s) - V^2(s)\\
    &= \pi Q^2(s) - 2R\pi Q(s) + 2RV(s) - V^2(s)\\
    &= \pi\roundy{Q^2 - 2RQ}(s) + 2RV(s) - V^2(s)\\
\end{align*}
Thus, for every $h$:
\begin{align*}
    \sum_{s\in \S_h}q(s)\var_{a\sim\pi}\roundy{Q(s,a)} &= \sum_{s\in \S_h}q(s)\roundy{2RV(s) - V^2(s)} + \sum_{s\in \S_h,a}q(s,a)\roundy{Q^2(s,a) - 2RQ(s,a)}
\end{align*}

Additionally, for every $s,a$:
\begin{align*}
    \var_{s'\sim P(s,a)}\roundy{V(s')} &= PV^2(s,a) - \roundy{PV(s,a)}^2\\
    &= PV^2(s,a) - Q^2(s,a) + Q^2(s,a) - \roundy{PV(s,a)}^2\\
    &= PV^2(s,a) - Q^2(s,a) + \roundy{Q(s,a) + PV(s,a)}\roundy{Q(s,a) - PV(s,a)}\\
    &\le PV^2(s,a) - Q^2(s,a) + 2RQ(s,a) - 2RPV(s,a)\\
    &= P(V^2 - 2RV)(s,a) + 2RQ(s,a) - Q^2(s,a)
\end{align*}

Thus, for every $h$:
\begin{align*}
    \sum_{s\in \S_h,a}q(s,a)\var_{s'\sim P(s,a)}\roundy{V(s')} &\le \sum_{s\in \S_{h+1}}q(s)\roundy{V^2(s) - 2RV(s)} + \sum_{s\in \S_h,a}q(s,a)\roundy{2RQ(s,a) - Q^2(s,a)}
\end{align*}

Summing both terms, for every $h$:
\begin{align*}
    \sum_{s\in \S_h,a}q(s,a)\var_{s'\sim P(s,a)}\roundy{V(s')} + &\sum_{s\in \S_h}q(s)\var_{a\sim\pi}\roundy{Q(s,a)} \le \\
    & \sum_{s\in \S_h}q(s)\roundy{2RV(s) - V^2(s)} - \sum_{s\in \S_{h+1}}q(s)\roundy{2RV(s) - V^2(s)}
\end{align*}

Telescoping for all $h$:
\begin{align*}
    \sum_{s,a}q(s,a)\var_{s'\sim P(s,a)}\roundy{V(s')} + \sum_{s}q(s)\var_{a\sim\pi}\roundy{Q(s,a)} &\le 2RV(s_1) - V^2(s_1)\le 2RV(s_1)
\end{align*}
\end{proof}
\begin{remark}
$Q$ doesn't need to to satisfy the Bellman equations exactly. One can see that the proof works as long as for all $s$, $V(s) = \pi Q(s)$, and for all $s,a$, $Q(s,a) - PV(s,a) \ge 0$ and $Q(s,a) \le R$.
\end{remark}

\begin{lemma}\label{lem:ltv}
For every $\pi$ and set of rewards $\curly{r(s,a)}_{s,a}$:
\begin{align*}
\sum_{s}q^\pi(s)\var_{a\sim \pi(s)}(Q^\pi(s,a)) + \sum_{s,a}q^\pi(s,a)\var_{s' \sim P(s,a)}(V^\pi(s')) &= \var_{\curly{s_h,a_h}_h \sim \pi}\roundy{\sum_{h=1}^H r(s_h,a_h)}\\
&\le H\sum_{s,a}q^\pi(s,a)r^2(s,a)
\end{align*}
\end{lemma}
\begin{proof}
Let $G_h = \sum_{h'=h}^H r(s_h,a_h)$. From the law of total variance:
\begin{align*}
    \var(G_h \mid s_h) &= \E_{a_h\sim \pi(s_h)}[\var(G_h \mid s_h, a_h)] + \var_{a_h \sim \pi(s_h)}(\E[G_h \mid s_h,a_h])\\
    &= \sum_{a_h}\pi(a_h \mid s_h)\var(G_h \mid s_h, a_h) + \var_{a_h \sim \pi(s_h)}(Q^\pi(s_h, a_h))
\end{align*}

Thus:
\begin{align*}
    \sum_{s_h}q^\pi(s_h)\var(G_h \mid s_h) &= \sum_{s_h,a_h}q^\pi(s_h,a_h)\var(G_h \mid s_h, a_h) + \sum_{s_h}q^\pi(s_h)\var_{a_h \sim \pi(s_h)}(Q^\pi(s_h, a_h))
\end{align*}

Using the law of total variance again:
\begin{align*}
    \var(G_{h+1}\mid s_h, a_h) &= \E_{s_{h+1}\sim P(s_h,a_h)}[\var(G_{h+1} \mid s_{h+1})] + \var_{s_{h+1} \sim P(s_h,a_h)}(\E[G_{h+1} \mid s_{h+1}])\\
    &= \sum_{s_{h+1}}P(s_{h+1}\mid s_h,a_h)\var(G_{h+1} \mid s_{h+1}) +  \var_{s_{h+1} \sim P(s_h,a_h)}(V^\pi(s_{h+1})
\end{align*}

Thus:
\begin{align*}
    \sum_{s_{h},a_h}q^\pi(s_h,a_h)\var(G_{h+1}\mid s_h, a_h) &= \sum_{s_{h+1}}q^\pi(s_{h+1})\var(G_{h+1} \mid s_{h+1}) +  \sum_{s_h,a_h}q^\pi(s_h,a_h)\var_{s_{h+1} \sim P(s_h,a_h)}(V^\pi(s_{h+1})
\end{align*}

Notice that since $G_h = r(s_h, a_h) + G_{h+1}$, and $r(s_h, a_h)$ is deterministic given $s_h,a_h$ we have:
\begin{align*}
    \var(G_h \mid s_h, a_h) = \var(G_{h+1}\mid s_h, a_h).
\end{align*}

Placing it back we get:
\begin{align*}
    \sum_{s_h}q^\pi(s_h)\var(G_h \mid s_h) = \sum_{s_{h+1}}q^\pi(s_{h+1})\var(G_{h+1} \mid s_{h+1}) &+ \sum_{s_h}q^\pi(s_h)\var_{a_h \sim \pi(s_h)}(Q^\pi(s_h, a_h)) \\
    &+ \sum_{s_h,a_h}q^\pi(s_h,a_h)\var_{s_{h+1} \sim P(s_h,a_h)}(V^\pi(s_{h+1})
\end{align*}

We have a sequence of the structure $a_h = a_{h+1} + b_h$, which means $a_1 = \sum_h b_h$, thus:
\begin{align*}
    \var(G_1) &= \sum_{s}q^\pi(s)\var_{a \sim \pi(s)}(Q^\pi(s, a)) + \sum_{s,a}q^\pi(s,a)\var_{s' \sim P(s,a)}(V^\pi(s'))
\end{align*}
\end{proof}

\subsection{Change of dynamics}
\begin{lemma}\label{lem:induction}
Let $V$ be the V-function w.r.t to rewards $r$ with policy $\pi$ and dynamics $P$, $\tilde{Q}(s,a) = \roundy{1 + \frac{1}{H}}P\tilde{V}(s,a) + r(s,a)$ for every $s,a$, where $\tilde{V}(s)=\sum_a \pi(a\mid s)\tilde{Q}(s,a)$. we have $\tilde{V} \le eV$.
\end{lemma}
\begin{proof}
We'll prove by induction:
\begin{align*}
    \tilde{V}(s) \le \roundy{1 + \frac{1}{H}}^{H-h(s)}V(s)
\end{align*}

The base is the fact that in $H+1$ both are $0$. For the step:
\begin{align*}
    \tilde{Q}(s,a) &\le \roundy{1 + \frac{1}{H}}P\roundy{\roundy{1 + \frac{1}{H}}^{H-h(s)-1}V}(s,a) + r(s,a)\\
    &= \roundy{1 + \frac{1}{H}}^{H-h(s)}PV(s,a) + r(s,a)\\
    &\le \roundy{1 + \frac{1}{H}}^{H-h(s)}\roundy{r(s,a) + PV(s,a)}\\
    &\le \roundy{1 + \frac{1}{H}}^{H-h(s)}Q(s,a).
\end{align*}
\end{proof}

\begin{definition}\label{lem:edyn}
For some $\delta \in (0,1)$, let $\calE_{dyn}^\delta$ be the event that:
\begin{align}\label{eq:goodevent}
    \abs{\bar{P}(s'\mid s,a) - P(s'\mid s,a)} \le \sqrt{\frac{6\log(SA/\delta)P(s'\mid s,a)}{\max\curly{N(s,a),1}}} + \frac{2\log(SA/\delta)}{\max\curly{N(s,a),1}}.
\end{align}
\end{definition}
\begin{lemma}\label{lem:edyn_prob}
$\calE_{dyn}^\delta$ is true w.p $1-\delta$.
\end{lemma}
\begin{proof}
From Bernstein inequality with union bound for every $s,a,s'$. We note that in the case of $N(s,a) = 0$ the above bound is greater than the trivial bound of $1$.
\end{proof}

\begin{lemma}\label{lem:change_of_measure}
Fix $r_1,r_2$ be two reward functions, let $\bar{P}$ an empirical dynamics using $\curly{N(s,a)}_{s,a}$ samples and Let $\pi$ be some policy. Denote $R\ge 1$ to be such that $\norm{r_1+r_2}_\infty \le R$. Denote:
\begin{align*}
    \alpha(s,a) = \max\curly{r_2(s,a) + \roundy{\bar{P} - P}V^\pi(s,a;\; r_1) + \frac{8H^2SR\log(SA/\delta)}{\max\curly{N(s,a),1}},\,0}.
\end{align*}

Then, assume $\calE_{dyn}^\delta$:
\begin{align*}
    \bar{V}^\pi(s_1;\; r_1+r_2) - V^\pi(s_1;\; r_1) \le eV^\pi(s_1;\; \alpha)
\end{align*}
where $V^\pi(s_1;\; \alpha)$ is the value function w.r.t $\alpha$.

The same proof works also for $V^\pi(s_1;\; r_1 + r_2) - \bar{V}^\pi(s_1;\; r_1) \le eV^\pi(s_1;\; \alpha)$.
\end{lemma}
\begin{proof}
Denote $\Delta(s) = \max\curly{\bar{V}^\pi(s) - V^\pi(s),\, 0}$. First, we have that:
\begin{align*}
    (\bP - P)\Delta(s,a) &\le  \sum_{s'}\abs{\bP(s'\mid s,a) - P(s'\mid s,a)}\Delta(s')\\
    &\le \sum_{s'}\roundy{\sqrt{\frac{6\log(SA/\delta)P(s'\mid s,a)}{\max\curly{N(s,a),1}}} + \frac{2\log(SA/\delta)}{\max\curly{N(s,a),1}}}\Delta(s')\\
    &\le \sum_{s'}\roundy{\frac{P(s'\mid s,a)}{H} + \frac{8H \log(SA/\delta)}{\max\curly{N(s,a),1}}}\Delta(s')\\
    &\le \frac{P\Delta(s,a)}{H} + \frac{8H^2SR\log(SA/\delta)}{\max\curly{N(s,a),1}}.
\end{align*}
The first inequality is absolute value, 
the second is \Cref{eq:goodevent}, the third is AM-GM, and the last is since $\Delta$ is bounded by $HR$.

If $\Delta(s)$ wasn't clipped we have:
\begin{align*}
    \Delta(s) &= \sum_a \pi(a)\roundy{r_2(s,a) + \bP\bVpi(s,a) - P\Vpi(s,a)}\\
    &\le \sum_a \pi(a)\roundy{r_2(s,a)+ P\Delta(s,a) + (\bP - P)\bVpi(s,a)}\\
    &\le \sum_a \pi(a)\roundy{r_2(s,a)+ P\Delta(s,a) + (\bP - P)\Delta(s,a) + (\bP - P)\Vpi(s,a)}\\
    &\le \sum_a \pi(a)\roundy{r_2(s,a)+ \roundy{1 + \frac{1}{H}}P\Delta(s,a) + (\bP - P)\bVpi(s,a) + \frac{8H^2SR\log(SA/\delta)}{\max\curly{N(s,a),1}}}.\\
    &\le \sum_a \pi(a)\roundy{\roundy{1 + \frac{1}{H}}P\Delta(s,a) + \alpha(s,a)}
\end{align*}
Since $\alpha \ge 0$, this is true also for the case that $\Delta(s) = 0$.

\Cref{lem:induction} concludes the proof.
\end{proof}

\subsection{Implicit Exploration in Rewards}
\begin{lemma}\label{lem:ix_helper}
For any $b$ and $a\in [0,1]$,
\begin{align*}
    e^{-ab} \le 1 - a + ae^{-b}
\end{align*}
\end{lemma}
\begin{proof}
Fix $b$ and define $f(a) = e^{-ab}$. We have $f''(a) = b^2e^{-ab} \ge 0$ which means that $f$ is convex. Thus:
\begin{align*}
    f(a) &\le (1-a)f(0) + af(1) = 1 - a + ae^{-b}
\end{align*}
\end{proof}

\begin{lemma}\label{lem:ix}
Let $X_k$ for $k \in [K]$ be an $\calF_k$ measurable indicator with $\E_k[X_k] = p_k$ for some $p_k \in [0,1]$. Let $1 > \gamma > 0$. Let $r_k \in [0,1]$ be $\calF_{k-1}$ measurable random variable, and let $\widehat{r}_k = \frac{X_k r_k + 3/2\gamma}{p_k + \gamma}$. Then, w.p $1-\delta$,
\begin{align*}
    \sum_k r_k &\le \sum_k \widehat{r}_k  + \frac{\log\roundy{\frac{1}{\delta}}}{\gamma}
\end{align*}
\end{lemma}
\begin{proof}
We have for every $k$:
\begin{align*}
    \E_k\squary{\exp\roundy{-\frac{\gamma X_k r_k}{p_k + \gamma}}} &\le \E_k\squary{1 - X_k + X_k\exp\roundy{-\frac{\gamma r_k}{p_k + \gamma}}}\\
    &= 1 + p_k\roundy{\exp\roundy{-\frac{\gamma r_k}{p_k + \gamma}} - 1}\\
    &\le 1 + p_k\roundy{-\frac{\gamma r_k}{p_k + \gamma} + \frac{1}{2}\frac{\gamma^2 r_k^2}{(p_k + \gamma)^2}}\\
    &\le 1 - \frac{\gamma p_k r_k}{p_k + \gamma} + \frac{1}{2}\frac{\gamma^2}{p_k + \gamma}\\
    &\le 1 - \frac{\gamma p_k r_k + \gamma^2r_k}{p_k + \gamma} + \frac{3}{2}\frac{\gamma^2}{p_k + \gamma}\\
    &= 1 - \gamma r_k + \frac{3}{2}\frac{\gamma^2}{p_k + \gamma}\\
    &\le \exp\roundy{-\gamma r_k + \frac{3}{2}\frac{\gamma^2}{p_k + \gamma}}
\end{align*}
The first inequality is \Cref{lem:ix_helper}, the second uses $e^{-x} \le 1 - x + \frac{1}{2}x^2$ which is true for $x \ge 0$, the third inequality uses $r_k \le 1$ and $p_k \le p_k + \gamma$, the fourth inequality uses again $r_k \le 1$, the fifth inequality uses $1 + x \le e^x$ which is true for $x\in \rr$.

 This establishes that:
 \begin{align*}
     \E_k[\exp(-\gamma \widehat{r}_k)] &\le \exp\roundy{-\gamma r_k + \frac{3}{2}\frac{\gamma^2}{p_k + \gamma}}\exp\roundy{-\frac{3}{2}\frac{\gamma^2}{p_k + \gamma}} =\exp(-\gamma r_k).
 \end{align*}

Thus, the process $Z_k = \exp\roundy{\sum_{t=1}^k \roundy{\gamma r_t - \gamma \widehat{r}_t}}$ is supermartingale w.r.t $F_{k}$:
\begin{align*}
    \E_k[Z_k] &= \E_k[\exp\roundy{\gamma r_k - \gamma \widehat{r}_k}]Z_{k-1} \le Z_{k-1}.
\end{align*}

Since $Z_0 = 1$, we have:
\begin{align*}
    \E\squary{\exp\roundy{\sum_{k=1}^K \roundy{\gamma r_k - \gamma \widehat{r}_k}}} \le 1.
\end{align*}

From Markov inequality, for every $\delta$
\begin{align*}
    \Pr\squary{\sum_k \roundy{\gamma r_k - \gamma \widehat{r}_k} > \log\roundy{\frac{1}{\delta}}} &\le \delta.
\end{align*}
\end{proof}

\subsection{Hedge}
\begin{lemma}\label{lem:mgf}
For every random variable $X\in [0,1]$ a.s:
\begin{align*}
    \ln \E\squary{e^{X}} &\le \E\squary{X} + \var\squary{X}
\end{align*}
\end{lemma}
\begin{proof}
Denote $Y \coloneqq X - \E[X]$. Since $Y\in [-1,1]$ a.s:
\begin{align*}
    e^Y &\le 1 + Y + Y^2\\
    \implies \E\squary{e^Y} &\le 1 + \var\squary{X}\\
    \implies \ln \E\squary{e^Y} &\le \var\squary{X}
\end{align*}
The first is since $e^x \le 1 + x + x^2$ for all $x\in [-1,1]$,
the second is because $\E[Y] = 0$ and $\E\squary{Y^2} = \var\squary{X}$ and the last is since $\ln(1+x) \le x$ for all $x > -1$.

Finally:
\begin{align*}
    \ln \E\squary{e^{X - \E[X]}} &= \ln \E\squary{e^{X}} - \E[X]\\
    \implies \ln \E\squary{e^{X}} &\le \E\squary{X} + \var\squary{X}
\end{align*}
\end{proof}

\begin{lemma}\label{lem:imp_hedge}
Let $\eta > 0$, and an arbitrary sequence $\{r_k \in [0,1/\eta]^d\}_{k=1}^{K}$.
Let $x_k \in \Delta_d$ be a sequence of vectors such that for all $a$, $x_1(a)=1/d$, and for all $k \in [K]$, $a \in [d]$,
\[
x_{k+1}(a)
=
\frac{x_k(a)e^{\eta r_k(a)}}
{\sum_{a' \in [d]} x_k(a')e^{\eta r_k(a')}}.
\]

Then for any $x^\star \in \Delta_d$,
\[
\sum_{k=1}^{K} \langle r_k, x^\star - x_k \rangle
\leq
\frac{\log d}{\eta}
+
\eta \sum_{k=1}^{K}\var_{a\sim x_k}\roundy{r_k(a)}
\]
\end{lemma}
% For every $s$, as long as $\eta \le \frac{1}{2H}$:
% \begin{align*}
%     \sum_{k,a} \roundy{\pi^\star(a\mid s) - \pi_k(a\mid s)}\bar Q^{\pi_k}(s,a) \le \frac{\log(N)}{\eta} + \sum_k \eta \var_{a\sim \pi_k}\squary{\bar Q^{\pi_k}(s,a)}
% \end{align*}
% \end{lemma}
\begin{proof}
Denote:
\begin{align*}
    W_k = \sum_a \exp\roundy{\sum_{k'=1}^{k-1}\eta r_{k'}(a)}
\end{align*}

We assume w.l.o.g that $x^\star$ is deterministic, and $a^\star$ is the deterministic action.

From standard analysis of Hedge (e.g \cite{hazan2016introduction}):
\begin{align*}
    \log\roundy{\frac{W_{k+1}}{W_{k}}} &= \log\E_{a\sim x_k}\squary{e^{\eta r_{k}(a)}}\\
    &\le \eta\E_{a\sim x_k}\squary {r_k(a)} + \eta^2\var_{a\sim x_k}\squary{r_k(a)}
\end{align*}
In the inequality we used \Cref{lem:mgf} (recall that $\eta r_k \le 1$ for all $k$).

Since $W_1 = d$, we have:
\begin{align*}
    \log W_{K+1} &\le \log(d) + \sum_k \eta\E_{a\sim x_k}\squary{r_k(a)} + \sum_k\eta^2\var_{a\sim x_k}\squary{r_k(a)}
\end{align*}

On the other hand:
\begin{align*}
    W_{K+1} \ge \exp\roundy{\sum_{k}\eta r_k(a^\star)}
\end{align*}

Thus:
\begin{align*}
    \sum_{k,a}\eta x^\star(a)r_k(a) &\le \log(d) + \sum_{k,a}\eta x_k(a)r_k(a) + \sum_k \eta^2\var_{a\sim x_k}\squary{r_k(a)}
\end{align*}

Rearranging and dividing by $\eta$ concludes the proof.
\end{proof}

\section{Generic Algorithm}
\begin{algorithm}[H]
\caption{VOPO: Variance Optimistic Policy Optimization}
\label{alg:generic}
\begin{algorithmic}[1]

\STATE \textbf{Input}: $\eta > 0$.

\STATE Initialize policy
$\pi_1(a\mid s)=1/A$ for all $s,a$.

\FOR{$k=1,\ldots,K$}

    \STATE Execute policy $\pi_k$ for one episode and observe
    rewards $r_k$.

    \STATE Set $\bar V_k(s)=0$ for all $s\in\mathcal S_{H+1}$.

    \FOR{$h=H,H-1,\ldots,1$}
        \FORALL{$s\in\mathcal S_h,\; a\in\mathcal A$}

            \STATE Compute
            \[
            Q_k(s,a)
            =
            r_k(s,a)
            +
            P
            V_k(s,a),
            \]
            \[
            B_k(s,a)
            =
            b_k(s,a)
            +
            P
            B_k(s,a),
            \]
            \[
            U_k(s,a) = Q_k(s,a) + B_k(s,a)
            \]
            where $b_k(s,a)$ is defined in
            \Cref{eq:bonus}.

        \ENDFOR

        \FORALL{$s\in\mathcal S_h$}
            \STATE Compute
            \[
             V_k(s)
            =
            \sum_{a\in\mathcal A}
            \pi_k(a\mid s)\,
             Q_k(s,a).
            \]
            \[
             B_k(s)
            =
            \sum_{a\in\mathcal A}
            \pi_k(a\mid s)\,
             B_k(s,a).
            \]
        \ENDFOR

    \ENDFOR

    \FORALL{$s\in\mathcal S,\; a\in\mathcal A$}
        \STATE Update policy
        \[
        \pi_{k+1}(a\mid s)
        =
        \frac{
        \pi_k(a\mid s)
        \exp\bigl(\eta  U_k(s,a)\bigr)
        }{
        \sum_{a'}
        \pi_k(a'\mid s)
        \exp\bigl(\eta U_k(s,a')\bigr)
        }.
        \]
    \ENDFOR

\ENDFOR

\end{algorithmic}
\end{algorithm}

We assume that the given $\eta$ and rewards are such that for every $k,s,a$:
\begin{align*}
    \eta r_k(s,a) \in [0, \frac{1}{12H}]
\end{align*}

\begin{align}\label{eq:bonus}
    b_k(s) &= \eta\var_{a\sim \pi_k(s)}\roundy{\Upik(s,a)}\\
    &= \eta\var_{a\sim \pi_k(s)}\roundy{\Qpik(s,a) + P\Bpik(s,a)}\notag
\end{align}
The second equation is since $b_k(s)$ is the same for every $a$ and thus doesn't affect the variance.

\begin{lemma}\label{lem:Qbounded}
For all $k,s,a$:
\begin{align*}
    \eta \Qpik(s,a) \le \frac{1}{12}
\end{align*}
\end{lemma}
\begin{proof}
\begin{align*}
     \eta \Qpik(s,a) &= \sum_{s',a'}\qpik(s',a'\mid s,a)\eta r_k(s',a')\\
     &\le \frac{1}{12H}\sum_{s',a'}\qpik(s',a'\mid s,a)\\
     &\le \frac{1}{12}
\end{align*}
Where the first inequality is from the assumption that $\eta r_k(s,a) \le \frac{1}{12H}$.
\end{proof}

\begin{lemma}\label{lem:Bbounded}
For all $k,s,a$:
\begin{align*}
    \eta\Bpik(s) &\le \frac{1}{12}\\
    \eta \Bpik(s,a) &\le \frac{1}{6}
\end{align*}
\end{lemma}
\begin{proof}
Fix $h$ Assume true for every $a$ and $s\in \S_{h'}$, $h' > h$. We have:
\begin{align*}
    \Bpik(s) &= \eta\sum_{s'}\qpik(s'\mid s)\var_{a'\sim \pi_k(s)}\roundy{\Qpik(s',a') + P\Bpik(s',a')}\\
    &\le 2\eta \frac{1}{4\eta}\roundy{\Vpik(s) + \Bpik(s)}\\
    &= \frac{1}{2}\roundy{\Vpik(s) + \Bpik(s)}\\
    \implies \Bpik(s) &\le \Vpik(s)
\end{align*}
The inequality uses \Cref{lem:var_bound} with $R=\frac{1}{4\eta}$ (which is true due to \Cref{lem:Qbounded} and the induction assumption). This proves the first part. For the second part:
\begin{align*}
    \eta b_k(s) &\le \eta \Bpik(s) \le \frac{1}{12}\\
    \eta P\Bpik(s,a) &\le \eta P\Vpik(s,a)\le \frac{1}{12}\\
    \eta \Bpik(s,a) &= \eta b_k(s) + \eta P\Bpik(s,a) \le \frac{1}{6}. 
\end{align*}
\end{proof}

\begin{lemma}\label{lem:B_base_bound}
For every $k$:
\begin{align*}
    \Bpik \le 6\eta H\sum_{s,a}\qpik(s,a)r_k^2(s,a)
\end{align*}
\end{lemma}
\begin{proof}
\begin{align*}
    \Bpik &\le 2\eta\sum_{s}\qpik(s)\roundy{\var_{a\sim \pi_k(s)}\roundy{\Qpik(s,a)} + \var_{a\sim \pi_k(s)}\roundy{\Bpik(s,a)}}\\
    &\le 2\eta H\sum_{s,a}\qpik(s,a)r_k^2(s,a) + \frac{2}{3}\Bpik\\
    \implies \frac{1}{3}\Bpik &\le 2\eta H\sum_{s,a}\qpik(s,a)r_k^2(s,a)\\
    \implies \Bpik &\le 6\eta H\sum_{s,a}\qpik(s,a)r_k^2(s,a)
\end{align*}
In the first inequality we used the fact that $\var(X+Y) \le 2\var(X) + 2\var(Y)$. For the first term of the second inequality we used \Cref{lem:ltv} and for the second term we used \Cref{lem:var_bound} with $R=\frac{1}{6\eta}$ (due to \Cref{lem:Bbounded}).
\end{proof}

\begin{lemma}\label{lem:generic_regret}
\begin{align*}
    \sum_k \Vpistar - \Vpik &\le \frac{H\log(A)}{\eta} + 6\eta H\sum_{k,s,a}\qpik(s,a)r_k^2(s,a)
\end{align*}
\end{lemma}
\begin{proof}
From \Cref{lem:vdl}, for every $k$:
\begin{align*}
    \Upistar - \Upik &= \sum_{s}\qpistar(s)\sum_a\roundy{\pi^\star(a\mid s) - \pi_k(a\mid s)}\Upik(s,a)
\end{align*}

We note that from \Cref{lem:Qbounded,lem:Bbounded}, $\eta \Upik(s,a) \le 1$, thus we can use \Cref{lem:imp_hedge}, which means:
\begin{align*}
    \sum_k \Upistar - \Upik &\le \sum_{s}\qpistar(s)\roundy{\frac{\log(A)}{\eta} + \sum_k b_k(s)}\\
    &= \frac{H\log(A)}{\eta} + \sum_k \Bpistar
\end{align*}
In the second we used the fact that $\sum_s \qpistar(s) = H$ and that by definition $\sum_s \qpistar(s) b_k(s) = \Bpistar$.

Thus:
\begin{align*}
    \sum_k \Vpistar - \Vpik &= \sum_k \Upistar - \Bpistar - \Upik + \Bpik\\
    &\le \frac{H\log(A)}{\eta} + \sum_k \Bpik\\
\end{align*}
\Cref{lem:B_base_bound} concludes the proof.
\end{proof}

\section{Full Info, Known Dynamics}
\subsection{Upper Bound}
\begin{theorem}\label{thm:full_info_known_dynamics}
In the full info regime, \Cref{alg:generic} with $\eta^{-1} = \sqrt{\frac{HK}{\log(A)}} + 12H$ has:
\begin{align*}
    \sum_k \Vpistar - \Vpik &= O(\sqrt{H^3K\log(A)})
\end{align*}
\end{theorem}
\begin{proof}
Since $\eta \le \frac{1}{12H}$ and $r_k(s,a) \le 1$ for all $k,s,a$, we can use \Cref{lem:generic_regret}. Thus:
\begin{align*}
    \sum_k \Vpistar - \Vpik &\le \frac{H\log(A)}{\eta} + 6\eta H\sum_{k,s,a}\qpik(s,a)\\
    &\le \frac{H\log(A)}{\eta} + 6\eta H^2K
\end{align*}

Placing $\eta$ we get:
\begin{align*}
    \sum_k \Vpistar - \Vpik &= O\roundy{\sqrt{H^3K\log(A)} + H^2\log(A)}
\end{align*}

If $H\log(A) \le K$, the first term dominates, which complete the proof. If $K \le H\log(A)$, then the trivial regret bound gives us,
\begin{align*}
    \sum_k \Vpistar - \Vpik  \le HK  \le \sqrt{H^3K\log(A)}
\end{align*}
\end{proof}

\subsection{Lower Bound}
\begin{theorem}\label{thm:full-info-known-lower}
For any algorithm and $A\ge 2,H \ge 1,S\ge H, K\ge \min\curly{H\log(A),\,\log(AS/H)}$ there is an MDP for which the regret is $\Omega\roundy{\sqrt{H^2K\min\curly{H\log(A),\,\log(AS/H)}}}$.
\end{theorem}
\begin{proof}
First, if $A > S/H$ we have that $\log(AS/H) = \Omega\roundy{\log(A)}$ and thus
\begin{align*}
    \Omega\roundy{\sqrt{H^2K\min\curly{H\log(A),\,\log(AS/H)}}} = \Omega\roundy{\sqrt{H^2K\log(A)}}.
\end{align*}
This lower bound is immediate from a single-state expert problem with $H$-scaled rewards (e.g \cite{cesa2006prediction}).

Additionally, if $H=1$, the bound becomes:
\begin{align*}
    \Omega\roundy{\sqrt{H^2K\min\curly{H\log(A),\,\log(AS/H)}}} = \Omega\roundy{\sqrt{K\log(A)}}.
\end{align*}
Which is again standard expert lower bound.

Continue assuming $A \le S/H$ and $H\ge 2$. Denote $n = \floor{\min\curly{H/2,\,\log_A(S/H)}}$, from the assumptions $n\ge 1$.
We construct reduction to the experts problem with reward scale of $H - n \ge \frac{H}{2}$ and $A^{n}$ experts. From the standard lower bound of the experts problem we get:
\begin{align*}
    \Omega\roundy{\sqrt{H^2K\log\roundy{A^n}}} &= \Omega\roundy{\sqrt{H^2K\log(A)n}}\\
    &= \Omega\roundy{\sqrt{H^2K\log(A)\min\curly{H,\,\log_A(S/H)}}}\\
    &= \Omega\roundy{\sqrt{H^2K\min\curly{H\log(A),\,\log(AS/H)}}},
\end{align*}
as needed.

The MDP is full tree of depth $n+1$, which means that any combination of $A$ actions in the first $n$ steps reaches a unique state. This means $A^n \le S/H$ leaves. Mark those steps as $s_{n+1}^1,s_{n+1}^2,\dots s_{n+1}^{A^{n}}$. Afterwards, the leaf state persist- each leaf has a corresponding state in each of the next steps and deterministically transform there, i.e for all $i\in [1,A^{n}]$, $a\in [A]$ and $h\in [n+1,H-1]$, $P\roundy{s_{h+1}^{i} \mid s_h^i, a} = 1$. Note that the consruction doesn't use more than $S$ states - which step has a maximum of $S/H$ states so the total number of states doesn't exceed $S$.

For any reward in $[0,H-n]$ we can embed it in the last $H-n$ by dividing it by $H-n$ in each step. Thus, the above construction means that for each of the $A^{n}$ possible vectors of actions in the first $n$ steps,  there is a unique reward in $[0,H-n]$, which concludes the reduction.
\end{proof}

\section{Bandit Info, Known Dynamics}\label{apx:bandit_known}
We use the rewards,
\begin{align*}
    \widehat{r}_k(s,a) = \frac{\mathds{1}_k[s,a]r_k(s,a) + 2\gamma}{\qpik(s,a) + \gamma},
\end{align*}
with $\gamma = 36H\eta$ and $\eta ^{-1} = \sqrt{\frac{SAK}{\log(SA/\delta)}} + 36H$

\begin{lemma}\label{lem:bandit_good_event}
For every $s,a$, w.p $1-\delta$:
\begin{align*}
    \sum_k r_k(s,a) &\le \sum_k \widehat{r}_k(s,a) + \frac{\log(SA/\delta)}{\gamma}
\end{align*}
\end{lemma}
\begin{proof}
Directly from \Cref{lem:ix} (we used $+ 2\gamma$ factor instead of $3/2$ which only increases $\widehat{r}_k$) with $r_k = r_k(s,a)$ and union bound for all $s,a$.
\end{proof}

\begin{lemma}\label{lem:rhat_bound}
\begin{align*}
    \eta \hat{r}_k(s,a) \le \frac{1}{12H}
\end{align*}
\end{lemma}
\begin{proof}
For every $k,s,a$:
\begin{align*}
    \hat{r}_k(s,a) = \frac{\mathds{1}_k[s,a]r_k(s,a) + 2\gamma}{\qpik(s,a) + \gamma} \le \frac{3}{\gamma} = \frac{1}{12H\eta}
\end{align*}
We used the fact that $\eta^{-1} \ge 36H$ and thus $\gamma \le 1$.
\end{proof}

\begin{lemma}\label{lem:bandit_good_event2}
With probability $1-\delta$:
\begin{align*}
    \sum_{k,s,a} \widehat{r}_k(s,a) &\le \sum_{k,s,a} 2\E_k[\widehat{r}_k(s,a)] + \frac{SA\log(1/\delta)}{3H\eta}
\end{align*}
\end{lemma}
\begin{proof}
Directly from \Cref{lem:freedman_consequence} with $X_t = \sum_{s,a} \widehat{r}_k(s,a)$ and $R = \frac{SA}{12H\eta}$ (we use \Cref{lem:rhat_bound}).
\end{proof}

\begin{lemma}\label{lem:bandit_good_event3}
With probability $1-\delta$:
\begin{align*}
    \sum_k \hVpik &\le \sum_{k, s,a}\qpik(s,a)\E_k [\widehat{r}_k(s,a)] + \roundy{H + 2\gamma SA}\sqrt{2K\log(1/\delta)}
\end{align*}
\end{lemma}
\begin{proof}
We have:
\begin{align*}
    \qpik(s,a)\widehat{r}_k(s,a) &\le \mathds{1}_k[s,a]r_k(s,a) + 2\gamma\\
    \implies \sum_{s,a}\qpik(s,a)\widehat{r}_k(s,a) &\le H + 2\gamma SA
\end{align*}

Thus, Azuma inequality concludes the proof.
\end{proof}

\begin{theorem}\label{thm:bandit_known_dynamics}
\Cref{alg:generic} with $r_k = \hat{r}_k$, w.p $1-3\delta$:
\begin{align*}
    \sum_k \Vpistar - \Vpik &= O\roundy{\sqrt{\log(SA/\delta)H^2SAK}}
\end{align*}
\end{theorem}
\begin{proof}
We assume throughout this proof that \Cref{lem:bandit_good_event,lem:bandit_good_event2,lem:bandit_good_event3} is true, which happens w.p $1-3\delta$.

Denote $\widehat{V}$ to be the value function w.r.t $\widehat{r}$.
From \Cref{lem:generic_regret}:
\begin{align*}
    \sum_k \hVpistar - \hVpik &\le \frac{H\log(A)}{\eta} + 6\eta H\sum_{k,s,a}\qpik(s,a)\hat{r}_k^2(s,a)\\
    &\le \frac{H\log(A)}{\eta} + 24\eta H \sum_{k,s,a} \widehat{r}_k(s,a)\\
    &\le \frac{H\log(A)}{\eta} + 24\eta H \roundy{\sum_{k,s,a}2\frac{\qpik(s,a)r_k(s,a) + 2\gamma}{\qpik(s,a) + \gamma} + \frac{SA\log\roundy{1/\delta}}{3H\eta}}\\
    &= O\roundy{\frac{H\log(A)}{\eta} + \eta HSAK + SA\log(1/\delta)}
\end{align*}
The second is since $\qpik(s,a) \widehat{r}_k(s,a) \le 4$ for all $k,s,a$, the third is from \Cref{lem:bandit_good_event2} and in the last we used $r_k(s,a) \le 1$ for all $k,s,a$ and $H\eta  = O(\gamma)$.

We have:
\begin{align*}
    \sum_{s,a}\qpik(s,a)\E_k[\hat{r}_k(s,a)]
    &= \sum_{s,a}\qpik(s,a)\roundy{\frac{\qpik(s,a)r_k(s,a)}{\qpik(s,a) + \gamma} + \frac{2\gamma}{\qpik(s,a) + \gamma}}\\
    &\le \sum_{s,a}\qpik(s,a)r_k(s,a) + 2\sum_{s,a}\gamma\\
    &= \Vpik + O\roundy{SAH\eta}
\end{align*}

Which means, from \Cref{lem:bandit_good_event3},
\begin{align*}
    \sum_k \hVpik &\le \sum_k \Vpik + O\roundy{SAK\gamma + \roundy{H + SAH\eta}\sqrt{K\log(1/\delta)}}
\end{align*}

And in the other hand:
\begin{align*}
    \sum_k \Vpistar &= \sum_{k,s,a}q^\star(s,a)r_k(s,a) \\
    &\le \sum_{s,a}q^\star(s,a)\roundy{\sum_k \hat{r}_k(s,a) + \frac{\log(SA/\delta)}{\gamma}}\\
    &= \sum_k\hVpistar + \frac{H\log(SA/\delta)}{\gamma}\\
    &= \sum_k\hVpistar + O\roundy{\frac{\log(SA/\delta)}{\eta}}
\end{align*}

Thus:
\begin{align*}
    \sum_k \Vpistar - \Vpik &\le \sum_k \hVpistar - \hVpik + O\roundy{SAKH\eta + \roundy{H + SAH\eta}\sqrt{K\log(1/\delta)} + \frac{\log(SA/\delta)}{\eta}}\\
    &\le O\roundy{SAKH\eta + \roundy{H + SAH\eta}\sqrt{K\log(1/\delta)} + \frac{H\log(SA/\delta)}{\eta} + SA\log(1/\delta)}
\end{align*}

Placing $\eta$ we get:
\begin{align*}
    \sum_k \Vpistar - \Vpik &= O\roundy{\sqrt{\log(SA/\delta)H^2SAK} + (H^2 + SA)\log(SA/\delta)}
\end{align*}

If $(H + \frac{SA}{H})\log(SA/\delta) \le K$, we have:
\begin{align*}
    (H^2 + SA)\log(SA/\delta) &\le H\sqrt{\roundy{H + \frac{SA}{H}}K}\\
    &\le \sqrt{\roundy{H^3 + SAH}K}\\
    &\le O\roundy{\sqrt{H^2SAK}}
\end{align*}
In the last we used the fact that $H \le S$.

 If $K \le \roundy{H + \frac{SA}{H}}\log(SA/\delta)$, then the trivial regret bound gives us,
\begin{align*}
    \sum_k \Vpistar - \Vpik  \le HK  \le \sqrt{\roundy{H^3 + SAH}K\log(SA/\delta)} \le O\roundy{\sqrt{H^2SAK\log(SA/\delta)}}
\end{align*}
\end{proof}

\section{Full Info, Unknown Dynamics}
\begin{algorithm}[h!]
\caption{Reduction from Known to Unknown Dynamics}
\label{alg:known-to-unknown}
\begin{algorithmic}[1]
\STATE Initialize epoch $i \gets 1$
\STATE Initialize sample counts $N_i(s,a) \gets 0$ for all $(s,a)$
\STATE Initialize empirical dynamics $P_i(s'\mid s,a) \gets 1/S$ for all $(s,a,s')$
\STATE Initialize the inner algorithm $\mathcal{A}$
\FOR{episode $k = 1,2,\ldots,K$}
    \STATE Run the inner algorithm $\mathcal{A}$ for episode $k$, using $\bar P_i$ as the known dynamics and $r_k + b_i$ as rewards.
    \STATE Execute the resulting policy in the true MDP
    \STATE Observe the trajectory of episode $k$ and update the sample counts $N_i(s,a)$
    \IF{$\exists (s,a)$ such that the number of samples of $(s,a)$ has doubled}
        \STATE $i \gets i+1$
        \STATE Set $N_i(s',a') \gets N_{i-1}(s',a')$ for all $s'\ne s,\,a'\ne a$
        \STATE Set $N_i(s,a) \gets 2N_{i-1}(s,a)$
        \STATE Construct $\bar P_i$ from $N_i$ samples.
        \STATE Restart the inner algorithm $\mathcal{A}$.
    \ENDIF
\ENDFOR
\end{algorithmic}
\end{algorithm}

The bonuses are:
\begin{align*}
    b_i(s,a) = \min\curly{\frac{\beta}{\max\curly{N_i(s,a), 1}},\, H},
\end{align*}
with $\beta$ to be determined below.

\newcommand{\Nik}{N_{i_k}}
\newcommand{\bPik}{\bar{P}_{i_k}}

\subsection{Good Event}
\begin{definition}
Let $\calE_1$ be the event that for every $k$, $\bPik$ fulfills $\calE_{dyn}^{1/K^2}$ (\Cref{lem:edyn}).
\end{definition}
\begin{lemma}
$\calE_1$ is true w.p $1-\frac{1}{K}$.
\end{lemma}
\begin{proof}
From \Cref{lem:edyn_prob} with union bound for all $k$.
\end{proof}

\begin{definition}
Let $\calE_2$ be the event that, for every $s,a,k$, w.p $1-\frac{1}{K}$:
\begin{align*}
    \abs{\roundy{\bPik - P}\Vpistar(s,a)} \le \sqrt{\frac{6\log(SAK)\var_{s'\sim P(s,a)}(\Vpistar(s'))}{\max\curly{\Nik(s,a), 1}}} + \frac{2H\log(SAK)}{\max\curly{\Nik(s,a), 1}}
\end{align*}
\end{definition}
\begin{lemma}
$\calE_2$ is true w.p $1-\frac{1}{K}$
\end{lemma}
\begin{proof}
From Bernstein inequality with union bound for all $s,a,k$. We note that in the case of $\Nik(s,a) = 0$ the above bound is greater than the trivial bound of $H$.
\end{proof}

\begin{definition}
Let $\calE_3$ be the event that:
\begin{align*}
    \sum_{k,s,a} \frac{q_k(s,a)}{\max\curly{\Nik(s,a), 1}} \le 2\sum_{k,s,a} \frac{\mathds{I}_k[s,a]}{\max\curly{\Nik(s,a), 1}} + 16\log(K)
\end{align*}
\end{definition}
\begin{lemma}
$\calE_3$ is true w.p $1-\frac{1}{K}$.
\end{lemma}
\begin{proof}
Directly from \Cref{lem:freedman_consequence}.
\end{proof}

\begin{definition}
Let $\calE$ be the intersection of $\calE_1,\calE_2,\calE_3$.
\end{definition}
\subsection{Analysis}

\begin{lemma}\label{lem:bonus}
Assume $\calE$, for every $k$:
\begin{align*}
    \sum_k \bBpik &\le 76SA\beta \log(K)
\end{align*}
\end{lemma}
\begin{proof}
\Cref{lem:change_of_measure} with $r_1 = 0$, $r_2 = b_k$ and $R=H$:
\begin{align*}
    \sum_k  \bBpik &\le e\sum_{k,s,a}\qpik(s,a)\roundy{b_k(s,a) + \frac{8H^3S\log(SAK)}{
    \max\curly{\Nik(s,a), 1}}}\\
    &\le \sum_{k,s,a}\qpik(s,a)\frac{4\beta}{\max\curly{\Nik(s,a), 1}}\\
    &\le 76SA\beta \log(K).
\end{align*}
The second is since $\beta \ge 48H^3S\log(SAK)$ and the last is from $\calE_3$ (see \Cref{lem:harmonic}).
\end{proof}

\begin{lemma}\label{lem:quasi_opt}
Assume $\calE$, for every $k$:
\begin{align*}
    \Vpistar - \bVpistar - \bBpistar &\le \frac{6\log(SAK)H^2}{\beta}
\end{align*}
\end{lemma}
\begin{proof}
Denote $\Delta(s,a) = \max\curly{\Qpistar(s,a) - \bQpistar(s,a) - \bBpistar(s,a),\, 0}$ and $\Delta(s) = \sum_a \pi^\star(a\mid s)\Delta(s,a)$. Denote $C$ to be the $V/Q$ function with respect to the real dynamics, $\pi^\star$ and the rewards:
\begin{align*}
    c(s,a) = \frac{3\log(SAK)\var_{s'\sim P(s,a)}(\Vpistar(s'))}{\beta}
\end{align*}

We prove by backward induction that:
\begin{align*}
    \Delta(s,a) \le \roundy{2-\frac{h(s)}{H}}C(s,a).
\end{align*}

First, since $\Delta(s) = \sum_a \pi^\star(a\mid s)\Delta(s,a)$ and $C(s) = \sum_a \pi^\star(a\mid s)C(s,a)$ this also means $\Delta(s) \le \roundy{2-\frac{h(s)}{H}}C(s)$. Second, 
We have from \Cref{lem:ltv} that $C(s) \le \frac{3H^2\log(SAK)}{\beta}$ for every $s$, and since $\Delta(s) \le 2C(s)$ proving this induction concludes the proof.

Fix $s,a$. First, if $\Delta(s,a) = 0$ the non-negativity of $C$ concludes the proof. Continue assuming $\Delta(s,a) > 0$.

If $b_k(s,a) = H$, we have
\begin{align*}
    \Delta(s,a) &\le P\Vpistar(s,a) - \bPik\roundy{\bVpistar + \bBpistar}(s,a) - H \le 0
\end{align*}
The inequality is since $P\Vpistar(s,a) \le H$. The non-negativity of $C$ concludes the proof for this case.

Continue assuming $b_k(s,a) < H$. We have:
\begin{align*}
    \Delta(s,a) &= \roundy{P - \bPik}\Vpistar + \bPik\roundy{\Vpistar - \bVpistar - \bBpistar}(s,a) - \frac{\beta}{\max\curly{\Nik(s,a), 1}}\\
    &\le \roundy{P - \bPik}\Vpistar + \bPik\Delta(s,a) - \frac{\beta}{\max\curly{\Nik(s,a), 1}}\\
    &=  P\Delta(s,a) + \roundy{\bPik - P}\Delta(s,a) + \roundy{P - \bPik}\Vpistar(s,a) - \frac{\beta}{\max\curly{\Nik(s,a), 1}}\\
\end{align*}

We'll bound the terms separately. First:
\begin{align*}
    P\Delta(s,a) &\le P\roundy{2 - \frac{h(s) + 1}{H}}C(s,a)
\end{align*}

Second:
\begin{align*}
    \roundy{\bPik - P}\Delta(s,a) &\le 2\sum_{s'}\abs{P(s'\mid s,a) - \bPik(s'\mid s,a)}C(s')\\
    &\le \sum_{s'} \roundy{\sqrt{\frac{24\log(SAK)P(s'\mid s,a)}{\max\curly{\Nik(s,a), 1}}} + \frac{4\log(SAK)}{\max\curly{\Nik(s,a), 1}}}C(s')\\
    &\le \sum_{s'}\roundy{\frac{P(s'\mid s,a)}{H} + \frac{10H\log(SAK)}{\max\curly{\Nik(s,a), 1}}}C(s')\\
    &\le \frac{{PC(s,a)}}{H} + \frac{10H^3S\log(SAK)}{\beta\max\curly{\Nik(s,a), 1}}\\
    &\le \frac{{PC(s,a)}}{H} + \frac{H\log(SAK)}{\max\curly{\Nik(s,a), 1}}
\end{align*}
The first is from the induction assumption, the second is $\calE_1$, the third is AM-GM, the fourth is from the bound of $C$ and the last is since $\beta \ge 10H^2S$.

third:
\begin{align*}
    \roundy{\bPik - P}\Vpistar &\le \sqrt{\frac{6\log(SAK)\var_{s'\sim P(s,a)}(\Vpistar(s'))}{\max\curly{\Nik(s,a), 1}}} + \frac{2H\log(SAK)}{\max\curly{\Nik(s,a), 1}}\\
    &\le c_k(s,a) + \frac{\beta}{2\max\curly{\Nik(s,a), 1}} + \frac{2H\log(SAK)}{\max\curly{\Nik(s,a), 1}}\\
\end{align*}
The first is $\calE_2$ and the second is AM-GM.

Combining everything:
\begin{align*}
    \Delta(s,a) &\le c_k(s,a) + \roundy{2 - \frac{h(s)}{H}}PC(s,a) + \frac{\beta}{2\max\curly{\Nik(s,a), 1}} + \frac{3H\log(SAK)}{\max\curly{\Nik(s,a), 1}} - \frac{\beta}{\max\curly{\Nik(s,a), 1}} \\
    &\le c_k(s,a) + \roundy{2 - \frac{h(s)}{H}}PC(s,a)\\
    &\le \roundy{2 - \frac{h(s)}{H}}\roundy{c(s,a) + PC(s,a)}\\
    &= \roundy{2 - \frac{h(s)}{H}}C(s,a).
\end{align*}
We used the fact that $\beta \ge 6H\log(SAK)$.
\end{proof}

\begin{lemma}\label{lem:reg}
Assume the inner algorithm is \Cref{alg:generic},
\begin{align*}
    \sum_k \bVpistar + \bBpistar - \bVpik - \bBpik &\le \frac{HSA\log(K)\log(A)}{\eta} + 12\eta H^2K + 76SA\beta \log(K)
\end{align*}
\end{lemma}
\begin{proof}
We first note that since each $s,a$ can have a maximum of $\log(K)$ epochs that starts because of them, there is a maximum of $\log(K)SA$ epochs. Let $K_i$ be the set of episodes for the $i$th epoch.

For each epoch $i$, from \Cref{lem:generic_regret}:
\begin{align*}
    \sum_{k\in K_i} \bVpistar + \bBpistar - \bVpik - \bBpik &\le \frac{H\log(A)}{\eta} + 6\eta H\sum_{k\in K_i,s,a}\bqpik(s,a)(r_k(s,a) + b_k(s,a))^2\\
    &\le \frac{H\log(A)}{\eta} + 12\eta H\sum_{k\in K_i,s,a}\bqpik(s,a)(r_k^2(s,a) + b_k^2(s,a))\\
    &\le \frac{H\log(A)}{\eta} + 12\eta H^2K_i + 12H\eta \sum_{k\in K_i,s,a}\bqpik(s,a)b_k^2(s,a)\\
    &\le \frac{H\log(A)}{\eta} + 12\eta H^2K_i + \sum_{k\in K_i}\bBpik
\end{align*}

Summing for all epochs:
\begin{align*}
    \sum_{k} \bVpistar + \bBpistar - \bVpik - \bBpik &\le \frac{HSA\log(K)\log(A)}{\eta} + 12\eta H^2K + \sum_{k}\bBpik
\end{align*}

From \Cref{lem:bonus}:
\begin{align*}
    \sum_{k} \bVpistar + \bBpistar - \bVpik - \bBpik &\le \frac{HSA\log(K)\log(A)}{\eta} + 12\eta H^2K + 76SA\beta \log(K)
\end{align*}
\end{proof}

\begin{theorem}\label{thm:full_info_unknown_dynamics}
Assume the inner algorithm is \Cref{alg:generic}, $\eta^{-1} = \sqrt{\frac{HK}{SA\log(A)\log(K)}} + 12H(H+1)$ and $\beta = \sqrt{\frac{H^2K}{SA}} + 48H^3S\log(SAK)$, we have:
\begin{align*}
    \sum_k \E[\Vpistar - \Vpik] &= O\roundy{\sqrt{H^2SAK\log^3(SAK)} + \sqrt{H^3SAK\log(A)\log(K)} + H^3S^2A\log^2(SAK)}
\end{align*}
\end{theorem}
\begin{proof}
We decompose the regret as:
\begin{align*}
    \sum_k \Vpistar - \Vpik &= \underbrace{\sum_k \Vpistar - \bVpistar - \bBpistar}_{\texttt{Quasi-Opt}} + \underbrace{\sum_k\bVpistar + \bBpistar - \bVpik - \bBpik}_{\texttt{Reg}} + \underbrace{\sum_k \bBpik}_{\texttt{Bonus}} + \underbrace{\sum_k \bVpik - \Vpik}_{\texttt{Bias}}
\end{align*}

Assume $\calE$ and the good event of \Cref{apx:profiles}, the terms are bounded in \Cref{lem:quasi_opt,lem:bonus,lem:reg,lem:change_of_measure_profiles}. We have:
\begin{align*}
    \sum_k \Vpistar - \Vpik &= O\big(\frac{\log(SAK)H^2K}{\beta} + SA\beta \log(K) + \frac{HSA\log(K)\log(A)}{\eta} + \eta H^2K +\\
    &\qquad +\sqrt{H^2SAK\log^3(SAK)} + H^2S^2A\log^2(SAK)\big)
\end{align*}

$\calE$ and the good event of \Cref{apx:profiles} are true w.p $1 - O\roundy{\frac{1}{K}}$. Thus, even if we have the trivial bound of $HK$ in the case they don't happen, it only adds $O(H)$, which doesn't affect the bound.

Placing $\beta,\eta$:
\begin{align*}
    \sum_k \E[\Vpistar - \Vpik] &= O\roundy{\sqrt{H^2SAK\log^3(SAK)} + \sqrt{H^3SAK\log(A)\log(K)} + H^3S^2A\log^2(SAK)}
\end{align*}
\end{proof}

\section{Profiles}\label{apx:profiles}

Fix $\pi_1,\dots,\pi_k$ such that for every $h$ and $s\in \S_h$, $\Vpik(s)$ is measurable from:
\begin{align*}
    \curly{N_{i_{k'}}(s',a')}_{k' \le k,s',a'} \cup \curly{\bar{P}_{i_{k'}}(\cdot \mid s',a')}_{k' \le k,a',s'\in \cup_{h'\ge h}S_{h'}}.
\end{align*}
where $N_{i_k}$ is the first $2^{\floor{\log_2 N_k(s,a)}}$ samples of each $s,a$ and $\bar{P}_{i_{k}}$ is the empirical dynamics from those samples.

We call each set of episodes that use the same dynamics "epoch". Denote $\bar{P}_i$ to be the dynamics of the $i$th epoch and $N_i$ to be the number of samples.

We denote $\bar{P}^l$ to be the empirical dynamics given the first $2^l$ samples of each $s,a$.

\subsection{Good Event}
\begin{definition}
Let $\calE_1$ be the event that for every $k$, $\bPik$ fulfills $\calE_{dyn}^{1/K^2}$ (\Cref{lem:edyn}).
\end{definition}
\begin{lemma}
$\calE_1$ is true w.p $1-\frac{1}{K}$.
\end{lemma}
\begin{proof}
From \Cref{lem:edyn_prob} with union bound for all $k$.
\end{proof}

\begin{definition}
Let $\calE_2$ be the event that, for every $l$ and a mapping $k(s,a)$, w.p $1-\frac{1}{K}$:
\begin{align*}
    \sum_{s,a}\max\curly{\roundy{\bar{P}^l - P}V_{k(s,a)}^{\pi_{k(s,a)}}(s,a),\, 0} \le \sqrt{\frac{32SA\sum_{s,a}\var_{s'\sim P(s,a)} V_{k(s,a)}^{\pi_{k(s,a)}}(s')\log^2(SAK)}{2^{l-1}}} + \frac{8SAH\log^2(SAK)}{2^{l-1}} 
\end{align*}
\end{definition}
\begin{lemma}
$\calE_2$ is true w.p $1-\frac{2}{K}$.
\end{lemma}
\begin{proof}
Fix $\curly{N_{i_{k'}}(s',a')}_{k',s',a'}$ and the mapping $k(s,a)$. We add to the mapping also the option to map to $K+1$, which means that $V_{k(s,a)} = 0$. By assumption, $V_k^{\pi_k}(s)$ for $s\in \S_h$ is independent of $\bar{P}$ for $h' < h$ for all $k$. 

and thus the sum in $E_2$ can be seen as martingale sum backwards over $h$. From \Cref{lem:freedman} with $c=H$ and $\epsilon=c^2$, w.p $1-2\delta$
\begin{align*}
    \sum_{s,a}\roundy{\bar{P}^l - P} V_{k(s,a)}^{\pi_{k(s,a)}}(s,a) \le \sqrt{\frac{8\sum_{s,a}\var_{s'\sim P(s,a)}V_{k(s,a)}^{\pi_{k(s,a)}}(s')\log(1/\delta)}{2^{l-1}}} + \frac{2H\log(1/\delta)}{2^{l-1}}.
\end{align*}

The cardinality of the set $\curly{N_{i_{k'}}(s',a')}_{k',s',a'}$ is bounded by $\roundy{4SAK}^{SA\log(K)+1}$ (see Lemma 5 in \cite{zhang2025settling}) and the cardinality of the mapping $k(s,a)$ is $(K+1)^{SA}$ ($K+1$ since we allow mapping to $K+1$), if we take
\begin{align*}
    \delta = \roundy{K\roundy{4SAK}^{SA\log(K)+1}(K+1)^{SA}}^{-1},
\end{align*}

union bound concludes the proof. We support the $\max\curly{\cdot,\, 0}$ because the mapping $k(s,a)$ can be $0$ as explained above.
\end{proof}

\begin{definition}
Let $\calE_3$ be the event that:
\begin{align*}
    \sum_{k,s,a} \frac{q_k(s,a)}{\max\curly{N_{i_k}(s,a),1}} \le 2\sum_{k,s,a} \frac{\mathds{I}_k[s,a]}{\max\curly{N_{i_k}(s,a),1}} + 16\log(K)
\end{align*}
\end{definition}
\begin{lemma}
$\calE_3$ is true w.p $1-\frac{1}{K}$.
\end{lemma}
\begin{proof}
Directly from \Cref{lem:freedman_consequence}.
\end{proof}

\begin{definition}
Let $\calE_4$ be the event that:
\begin{align*}
    \sum_{k,s,a}q^{\pi_k}(s,a)\max\curly{\roundy{\bar{P}_{i_k} - P}V_k^{\pi_k}(s,a),\,0} \le 2\sum_{k,h}\max\curly{\roundy{\bar{P}_{i_k} - P}V_k^{\pi_k}(s_k^h,a_k^h),\,0} + 16H\log\roundy{K}
\end{align*}
\end{definition}
\begin{lemma}
$\calE_4$ is true w.p $1-\frac{1}{K}$.
\end{lemma}
\begin{proof}
Directly from the second part of \Cref{lem:freedman_consequence}.
\end{proof}

\begin{definition}
Let $\calE_5$ be the event that:
\begin{align*}
    \sum_{k,h}\var_{s'\sim P(s_k^h,a_k^a)} V_{k}^{\pi_k}(s') \le 2\sum_{k,s,a}q^{\pi_k}(s,a)\var_{s'\sim P(s,a)} V_{k}^{\pi_k}(s') + 4H^2\log\roundy{K}
\end{align*}
\end{definition}
\begin{lemma}
$\calE_5$ is true w.p $1-\frac{1}{K}$.
\end{lemma}
\begin{proof}
Directly from the first part of \Cref{lem:freedman_consequence}.
\end{proof}

\subsection{Proof}
\begin{lemma}\label{lem:harmonic}
Assume $\calE_3$,
\begin{align*}
    \sum_{k,s,a}\frac{\qpik(s,a)}{\max\curly{N_{i_k}(s,a), 1}} &\le 19SA\log(K)
\end{align*}
\end{lemma}
\begin{proof}
From $\calE_3$:
\begin{align*}
    \sum_{k,s,a} \frac{q_k(s,a)}{\max\curly{N_{i_k}(s,a), 1}} &\le 2\sum_{k,s,a} \frac{\mathds{I}_k[s,a]}{\max\curly{N_{i_k}(s,a), 1}} + 16\log(K)\\
    &\le 2\sum_{s,a}(\log(K) + 1) + 16\log(K)\\
    &\le 19SA\log(K)
\end{align*}
The first inequality is due to the harmonic series bound.
\end{proof}

\begin{lemma}\label{lem:profiles}
Assume $\calE_2,\calE_4,\calE_5$:
\begin{align*}
    \sum_{k,s,a}q^{\pi_k}(s,a)\max\curly{\roundy{\bar{P}_{i_k} - P}V_k^{\pi_k}(s,a),\,0} = {O}\roundy{\sqrt{H^2KAS\log^3(SAK)}}
\end{align*}
\end{lemma}
\begin{proof}
Denote $ V_{k_{l,j,s,a}}$ to be ${V_k^{\pi_k}}$ in the episode of the $2^{l-1}+j$th visit of $s,a$. We have:
\begin{align*}
    \sum_{k,h}\max\curly{\roundy{\bar{P}_{i_k} - P} V_k^{\pi_k}(s_k^h,a_k^h),\,0}&= \sum_{l=0}^{\log(K)}\sum_{k,h,s\in\S_h,a}\max\curly{\roundy{\bar{P}^l - P}V_k^{\pi_k}(s_k^h,a_k^h)\indicator{N_{i_k}(s,a) = 2^{l-1}},\, 0}\\
    &\le \sum_{l=1}^{\log(K)}\sum_{k,h,s\in\S_h,a}\max\curly{\roundy{\bar{P}^l - P}V_k^{\pi_k}(s_k^h,a_k^h)\indicator{N_{i_k}(s,a) = 2^{l-1}}, \,0} + SAH\\
    &= \sum_{l=1}^{\log(K)}\sum_{j=1}^{2^{l-1}}\sum_{s,a}\max\curly{\roundy{\bar{P}^l - P}V_{k_{l,j,s,a}}(s,a),\,0} + SAH
\end{align*}

We now use $\calE_2$ to bound the above. Since $\calE_2$ is true for every mapping $k(s,a)$, it is true for the mapping $k(s,a) = k_{l,j,s,a}$ for every $l,j$. Thus:
\begin{align*}
    \sum_{k,h}\max\curly{\roundy{\bar{P}_{i_k} - P} V_k^{\pi_k}(s_k^h,a_k^h),\,0}  &\le \sum_{l=1}^{\log(K)}\sum_{j=1}^{2^{l-1}}\sqrt{\frac{32SA\sum_{s,a}\var_{s'\sim P(s,a)} V_{k_{l,j,s,a}}(s')\log^2(SAK)}{2^{l-1}}} + \frac{8SAH\log^2(SAK)}{2^{l-1}} \\
    &\le \sqrt{32SA\sum_{l=1}^{\log(K)}\sum_{j=1}^{2^{l-1}}\sum_{s,a}\var_{s'\sim P(s,a)} V_{k_{l,j,s,a}}(s')\log^3(SAK)} + 8SAH\log^3(SAK)\\
    &\le \sqrt{32SA\sum_{k,h}\var_{s'\sim P(s_k^h,a_k^a)} V_{k}^{\pi_k}(s')\log^3(SAK)} + 8SAH\log^3(SAK)
\end{align*}
The first inequality is Cauchy-Schwarz.

Applying $\calE_4$ and $\calE_5$:
\begin{align*}
    \sum_{k,s,a}&q^{\pi_k}(s,a)\max\curly{\roundy{\bar{P}_{i_k} - P}V_k^{\pi_k}(s,a),\,0} \\&\le 2\sum_{k,h}\max\curly{\roundy{\bar{P}_{i_k} - P} V_k^{\pi_k}(s_k^h,a_k^h),0} + 16H\log\roundy{K} \\
    &\le \sqrt{256SA\sum_{k,s,a}q^{\pi_k}(s,a)\var_{s'\sim P(s,a)} V_{k}^{\pi_k}(s')\log^3(SAK)} + 272SAH\log^3(SAK)\\
    &\le \sqrt{256SAKH^2\log^3(SAK)} + 272SAH\log^3(SAK),
\end{align*}

where the last is from \Cref{lem:ltv}.

Note that if $SA\log^3(SAK) \ge K$ the first term becomes trivial $HK$. Thus, the first term dominates.
\end{proof}

\begin{lemma}\label{lem:change_of_measure_profiles}
Assume $\calE$,
\begin{align*}
    \sum_k \bVpik - \Vpik =  O\roundy{\sqrt{H^2SAK\log^3(SAK)} + H^2S^2A\log^2(SAK)}.
\end{align*}
\end{lemma}
\begin{proof}
From \Cref{lem:change_of_measure} and $\calE_1$, for all $k$:
\begin{align*}
    \bVpik - \Vpik &\le e\sum_{s,a}\qpik(s,a)\roundy{\max\curly{(\bPik - P)\Vpik(s,a),\,0} + \frac{8H^2S\log(SAK)}{\max\curly{\Nik(s,a), 1}}}
\end{align*}

\Cref{lem:harmonic,lem:profiles} concludes the proof.
\end{proof}

\section{Useful existing results}
\begin{lemma}[Freedman's Inequality, Theorem 1 in \cite{beygelzimer2011contextual}]\label{lem:beygelzimer}
Let $\{X_t\}_{t \geq 1}$ be a real-valued martingale difference sequence adapted to a filtration $\{\mathcal{F}_t\}_{t \geq 0}$. If $|X_t| \leq R$ a.s., then for any $\eta \in (0,1/R]$, $T \in \mathbb{N}$, it holds with probability greater than $1-\delta$ that
\[
\sum_{t=1}^{T} X_t
\leq
\eta \sum_{t=1}^{T} \mathbb{E}[X_t^2 \mid \mathcal{F}_{t-1}]
+
\frac{\log(1/\delta)}{\eta}.
\]
\end{lemma}

\begin{lemma}[Freedman's inequality, Lemma 15 in \cite{zhang2025settling}]\label{lem:freedman}
Let $(M_n)_{n \geq 0}$ be a martingale such that $M_0 = 0$ and
$|M_n - M_{n-1}| \leq c$ $(\forall n \geq 1)$ hold for some quantity $c > 0$.
Define
\[
\operatorname{Var}_n := \sum_{k=1}^{n} \mathbb{E}\!\left[(M_k - M_{k-1})^2 \mid \mathcal{F}_{k-1}\right]
\]
for every $n > 0$, where $\mathcal{F}_k$ is the $\sigma$-algebra generated by
$(M_1,\ldots,M_k)$.
Then for every integer $n \geq 1$ and any $\delta > 0$, one has
\[
\mathbb{P}\!\left[
|M_n|
\geq
2\sqrt{2}\sqrt{\operatorname{Var}_n \log \frac{1}{\delta}}
+
2\sqrt{\epsilon \log \frac{1}{\delta}}
+
2c \log \frac{1}{\delta}
\right]
\leq
2\left(
\log_2 \left(\frac{nc^2}{\epsilon}\right) + 1
\right)\delta .
\]
\end{lemma}

\begin{lemma}[A slight modification from Lemma 27 in \cite{efroni2021confidence}]\label{lem:freedman_consequence}
Let $\{ X_t \}_{t\geq 1}$ be a sequence of random variables, supported in $[0,R]$, and adapted to a filtration $\calF_1 \subseteq \calF_2 \subseteq...\calF_T$. For any $T$, with probability $1-\delta$,
\begin{align*}
     \sum_{t=1}^T X_t &\leq 2 \sum_{t=1}^T\bbE[X_t \mid \calF_{t-1}] + 4R \log\roundy{\frac{1}{\delta}}\\
      \sum_{t=1}^T\bbE[X_t \mid \calF_{t-1}] &\leq 2 \sum_{t=1}^T X_t + 16R \log\roundy{\frac{1}{\delta}}
\end{align*}
\end{lemma}
\begin{proof}
We have for every $t$:
\begin{align*}
    \E\squary{\roundy{X_t - \bbE[X_t \mid \calF_{t-1}]}^2 \mid \calF_{t-1}} &\le 2\E\squary{{X_t}^2 \mid \calF_{t-1}} + 2\E\squary{{\bbE[X_t \mid \calF_{t-1}]}^2 \mid \calF_t} \tag{$a-b^2 \le 2a^2 + 2b^2$}\\
    &\le 2\E\squary{{X_t}^2 \mid \calF_{t-1}} + 2\E\squary{{\bbE[X_t^2 \mid \calF_{t-1}]} \mid \calF_t} \tag{Jensen's inequality}\\
    &\le 4\E\squary{{X_t}^2 \mid \calF_{t-1}} \tag{Tower rule}
    \\
    &\le 4R\,\E\squary{{X_t} \mid \calF_{t-1}}
\end{align*}

From \Cref{lem:beygelzimer} with $\eta = \frac{1}{4R}$:
\begin{align*}
    X_t - \bbE[X_t \mid \calF_{t-1}] &\le \E\squary{{X_t} \mid \calF_{t-1}} + 4R\log(1/\delta)\\
    \implies \sum_{t=1}^T X_t &\leq 2 \sum_{t=1}^T\bbE[X_t \mid \calF_{t-1}] + 4R \log\roundy{\frac{1}{\delta}}
\end{align*}

Again from \Cref{lem:beygelzimer}, now with $\eta = \frac{1}{8R}$:
\begin{align*}
    \bbE[X_t \mid \calF_{t-1}] - X_t &\le \frac{1}{2}\E\squary{{X_t} \mid \calF_{t-1}} + 8R\log(1/\delta)\\
    \implies \bbE[X_t \mid \calF_{t-1}] &\le 2X_t + 16R\log(1/\delta)\\
    \implies\sum_{t=1}^T\bbE[X_t \mid \calF_{t-1}] &\leq 2 \sum_{t=1}^T X_t + 16R \log\roundy{\frac{1}{\delta}}
\end{align*}
\end{proof}

\end{document}